\documentclass{article}

\usepackage[utf8]{inputenc} 
\usepackage[T1]{fontenc}    
\usepackage[dvipsnames]{xcolor}
\usepackage[english]{babel}
\usepackage{blindtext}
\usepackage{hyperref}
\definecolor{mydarkblue}{rgb}{0,0.08,0.45}
\hypersetup{ %
    colorlinks=true,
    linkcolor=mydarkblue,
    citecolor=mydarkblue,
    filecolor=mydarkblue,
    urlcolor=mydarkblue,
    pdfview=FitH}
    
\usepackage{url}            
\usepackage{booktabs}       
\usepackage{amsfonts}       
\usepackage{nicefrac}       
\usepackage{microtype}      
\usepackage{microtype}
\usepackage{graphicx}
\usepackage{booktabs} 
\usepackage{float}
\usepackage{multirow}
\usepackage{wrapfig}
\usepackage{tabularx}
\usepackage{tikz}
\usepackage{algpseudocode}
\usetikzlibrary{patterns}
\usetikzlibrary{positioning}

\algnewcommand{\Hyperparameters}[1]{%
  \State \textbf{hyperparameters:} #1
}
\algnewcommand{\Input}[1]{%
  \State \textbf{input:}
  \Statex \hspace*{\algorithmicindent}\parbox[t]{.8\linewidth}{\raggedright #1}
}
\algnewcommand{\Initialization}[1]{%
  \State \textbf{initialization:}
  \Statex \hspace*{\algorithmicindent}\parbox[t]{.8\linewidth}{\raggedright #1}
}
\algnewcommand{\Prerequisites}[1]{%
  \State \textbf{prerequisites:}
  \Statex \hspace*{\algorithmicindent}\parbox[t]{.8\linewidth}{\raggedright #1}
}
\algnewcommand{\Initialize}[1]{%
  \State \textbf{initialize} #1
}
\algnewcommand{\Notation}[1]{%
  \State \textbf{notation:} #1
}
\algnewcommand{\Note}[1]{%
  \State \textbf{note:} #1
}
\algblock[At]{At}{EndAt}
\algblockdefx[At]{At}{EndAt}[1]{\textbf{at} #1 \textbf{do}}{\textbf{end at}}

\usepackage{pifont}
\newcommand{\cmark}{\ding{51}}%
\newcommand{\xmark}{\ding{55}}%

\usepackage{hyperref}

\usepackage[final]{neurips_2019}

\newif\ifpaperchecker
\papercheckerfalse

\usepackage{amsfonts}
\usepackage{amsmath}
\usepackage{amsthm}
\usepackage{amssymb}
\usepackage{mathtools}
\usepackage{dsfont}
\usepackage[T1]{fontenc}
\usepackage{color}
\usepackage{graphicx}
\usepackage{wrapfig}
\usepackage{xparse}
\usepackage{enumitem}
\usepackage{multirow}
\usepackage{array}
\usepackage{blkarray}
\usepackage{bm}
\graphicspath{{figs/}}  

\usepackage{algorithm}
\usepackage{algpseudocode}
\usepackage{booktabs}
\usepackage{verbatim}
\usepackage{subcaption}

\DeclarePairedDelimiterX{\lin}[2]{\langle}{\rangle}{#1, #2}
\DeclarePairedDelimiterX{\abs}[1]{\lvert}{\rvert}{#1}
\DeclarePairedDelimiterX{\norm}[1]{\lVert}{\rVert}{#1}
\DeclarePairedDelimiterX{\cbr}[1]{\{}{\}}{#1} 
\DeclarePairedDelimiterX{\rbr}[1]{(}{)}{#1} 
\DeclarePairedDelimiterX{\sbr}[1]{[}{]}{#1} 

  \providecommand{\R}{\mathbb{R}} 
  \providecommand{\real}{\mathbb{R}} 

  \DeclareMathOperator{\expect}{\mathbb{E}}

  \DeclareMathOperator{\sgn}{sign}
  \makeatletter
  \def\sign{\@ifnextchar*{\@sgnargscaled}{\@ifnextchar[{\sgnargscaleas}{\@ifnextchar{\bgroup}{\@sgnarg}{\sgn} }}}
  \def\@sgnarg#1{\sgn\rbr{#1}}
  \def\@sgnargscaled#1{\sgn\rbr*{#1}}
  \def\@sgnargscaleas[#1]#2{\sgn\rbr[#1]{#2}}
  \makeatother

  \DeclareMathOperator*{\argmin}{arg\,min}

  \providecommand{\0}{\mathbf{0}}

  \providecommand{\ee}{\mathbf{e}}

  \renewcommand{\gg}{\mathbf{g}}

  \providecommand{\mm}{\mathbf{m}}

  \providecommand{\uu}{\mathbf{u}}
  \providecommand{\vv}{\mathbf{v}}
  
  \providecommand{\xx}{\mathbf{x}}

  \providecommand{\cC}{\mathcal{C}}

  \providecommand{\cO}{\mathcal{O}}

\newtheorem{theorem}{Theorem}

\newtheorem{lemma}{Lemma}
\newtheorem{remark}[lemma]{Remark}

\newtheorem{definition}{Definition}

\newtheorem{assumption}[definition]{Assumption}

\newcommand{\cifar}{\textsc{Cifar10}}
\newcommand{\wikitext}{\textsc{Wikitext-2}}
\newcommand{\resnet}{\textsc{ResNet18}}
\newcommand{\nccl}{\textsc{NCCL}}
\newcommand{\gloo}{\textsc{GLOO}}
\newcommand{\pytorch}{\textsc{PyTorch}}
\newcommand{\powersgd}{\textsc{PowerSGD}}
\newcommand{\atomo}{Spectral Atomo}
\newcommand{\signum}{Signum} 
\newcommand{\eg}{e.g.}
\newcommand{\ie}{i.e.}
\newcommand{\iid}{i.i.d.}
\newcommand{\cf}{cf.}
\newcommand{\allreduce}{all-reduce}
\newcommand{\allgather}{all-gather}

\definecolor{color1}{RGB}{228,26,28}
\definecolor{color2}{RGB}{55,126,184}
\definecolor{color3}{RGB}{77,175,74}
\definecolor{color4}{RGB}{152,78,163}
\definecolor{color5}{RGB}{255,127,0}

\newcommand{\tablefontsize}{\footnotesize}

\def\commentType{0}

\ifnum\commentType=0
  \usepackage[disable]{todonotes} 
  \newcommand{\customComment}[3]{}
  \newcommand{\customTODO}[3]{}
\fi
\ifnum\commentType=1
  \newcommand{\customComment}[3]{\textcolor{#2}{\textsl{#1:~#3}}}
  \newcommand{\customTODO}[3]{\textcolor{#2}{\textsl{#1:~#3}}}
\fi
\ifnum\commentType=2
  \usepackage{pdfcomment}
  \newcommand{\customComment}[3]{\pdfcomment[icon=Comment,opacity=0.5,color=#2,author=#1]{#3}}
  \newcommand{\customTODO}[3]{\pdfcomment[icon=Note,opacity=0.5,color=#2,author=#1]{#3}}
\fi
\ifnum\commentType=3
  \usepackage{todonotes} 
  \usepackage{silence}
  \newcommand{\customComment}[3]{\todo[color=#2!40,size=\small]{\textbf{#1:} #3}}
  \newcommand{\customTODO}[3]{\todo[inline,color=#2!40,size=\small]{\textbf{#1:} #3}}
\fi

\newcommand{\mj}[1]{\customComment{M}{green}{#1}}

\title{PowerSGD: Practical Low-Rank \\Gradient Compression for Distributed Optimization}
\author{%
  Thijs Vogels\\
  EPFL\\
  Lausanne, Switzerland \\
  \texttt{thijs.vogels@epfl.ch} \\
  \And
  Sai Praneeth Karimireddy \\
  EPFL\\
  Lausanne, Switzerland \\
  \texttt{sai.karimrieddy@epfl.ch} \\
   \And
  Martin Jaggi\\
  EPFL\\
  Lausanne, Switzerland \\
  \texttt{martin.jaggi@epfl.ch}
}

\begin{document}

\maketitle


\begin{abstract}
  We study lossy gradient compression methods to alleviate the communication bottleneck in data-parallel distributed optimization.
  Despite the significant attention received, current compression schemes either do not scale well, or fail to achieve the target test accuracy.
  We propose a new low-rank gradient compressor based on power iteration that can i)~compress gradients rapidly, ii)~efficiently aggregate the compressed gradients using \allreduce, and iii)~achieve test performance on par with SGD.
  The proposed algorithm is the only method evaluated that achieves consistent wall-clock speedups when benchmarked against regular SGD using highly optimized off-the-shelf tools for distributed communication.
  We demonstrate reduced training times for convolutional networks as well as LSTMs on common datasets.
  Our code is available at \url{https://github.com/epfml/powersgd}.
\end{abstract}

\section{Introduction}
Synchronous data-parallel SGD is the most common method for accelerating training of deep learning models \citep{dean2012large, iandola2015firecaffe, goyal2017accurate}.
Because the gradient vectors of such models can be large, the time required to share those gradients across workers limits the scalability of deep learning training~\citep{seide20141,iandola2015firecaffe, lin2017deep}.

Previous work proposes lossy gradient compression as a solution to this issue.
Notable examples include replacing the coordinates of the gradient with only their sign \citep{seide20141,Carlson:2015to,bernstein2018signsgd,bernstein2019iclr,karimireddy2019error}, quantizing the individual coordinates \citep{alistarh2017quantized,wen2017terngrad}, and low-rank approximation of the gradient \citep{wang2018atomo}.
While these works demonstrate speedups over full-precision SGD in some settings, we find that their speedups vanish with a fast network and highly optimized communication backend, even on commodity hardware.
Some prior work also suffers from degraded test accuracy compared to SGD.
We combine three observations to fix these issues: i) Linear compressor operators achieve scalability by enabling aggregation using \allreduce. ii) Error feedback ensures convergence with general biased compressors. iii) Low-rank updates enable aggressive compression without sacrificing quality.

First, we explore the properties of various gradient compression schemes for SGD and identify which ones are crucial for high scalability.
In particular, we note that currently proposed gradient compressors are not linear.
Their compressed messages cannot be added up hierarchically, unlike raw gradients.
This prevents current compressed SGD algorithms from aggregating gradients using an efficient \emph{reduce} operation and instead require a \emph{gather} operation.
Current deep learning frameworks rely either solely or predominantly on \allreduce, which is key to why regular SGD scales well with fast communication hardware~\citep[cf.][]{awan2018optimized, panda2019high}.

Secondly, it was recently shown that using error feedback (\ie\ storing the difference between the computed and compressed gradient, and reinserting it at the next iteration) improves both convergence and generalization for compression schemes \citep{karimireddy2019error}.
This can enable general biased gradient compression schemes to reach the target test accuracy.


Thirdly, there is growing evidence that the generalization ability of modern over-parameterized deep learning models is related to low-rankedness \citep{arora2018stronger, martin2018implicit,collins2018memorization}. Using a low-rank update (as we do) can be viewed as implicitly performing spectral regularization \citep{gunasekar2018characterizing} and hence can be expected to have good generalization properties \citep{yoshida2017spectral}.
Further, \cite{wang2018atomo} show that the eigenspectrum of the stochastic gradients for deep learning models decays, suggesting that a rank-based schemes can get away with aggressive compression without sacrificing convergence.

In this work, we design \powersgd\ with the above observations in mind. \powersgd\ computes a low-rank approximation of the gradient using a generalized \emph{power} iteration (known as subspace iteration \citep{stewart1975methods}).
The approximation is computationally light-weight, avoiding any prohibitively expensive Singular Value Decomposition.
To improve the quality of the efficient approximation, we \emph{warm-start} the power iteration by reusing the approximation from the previous optimization step.
Using \allreduce\ gradient aggregation, we empirically demonstrate that \powersgd\ achieves wall-clock speedups over regular SGD in a 16-GPU setting, even with the optimized \nccl\ communication backend on a fast network (and is the only algorithm to do so.)
By compressing gradients more than $120\times$, we reduce communication time (including coding and decoding) by $54\%$ for \resnet\ on \cifar\ and by $90\%$ for an LSTM on \wikitext.
End-to-end wall-clock training time to full test quality is reduced by $24\%$ for \resnet\ and by $55\%$ for the LSTM.

\section{Related work}

\begin{figure}[]
  \input{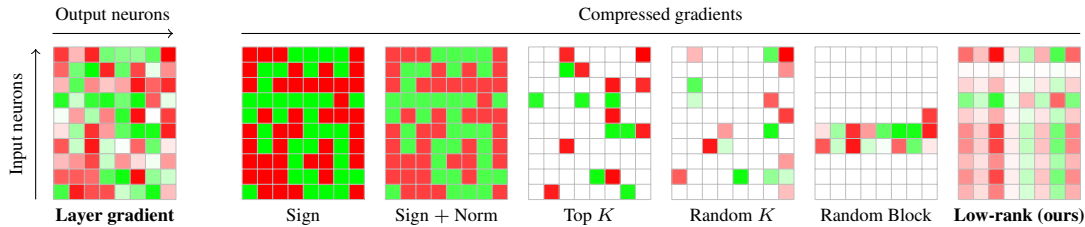}
  \caption{
    Compression schemes compared in this paper.
    Left: Interpretation of a layer's gradient as a matrix. 
    \ifpaperchecker\else
    Coordinate values are color coded ({\color{green} \textbf{positive}}, {\color{red} \textbf{negative}}). 
    \fi
    Right: The output of various compression schemes\ifpaperchecker\else on the same input\fi.
    Implementation details \ifpaperchecker\else are\fi in Appendix~\ref{sec:implementations}.}
    \ifpaperchecker\vspace{-4mm}\else\vspace{-3mm}\fi
  \label{fig:compression_methods}
\end{figure}

\paragraph{Gradient compression}
A variety of compression schemes (Figure~\ref{fig:compression_methods}) have been proposed: \cite{alistarh2017quantized} and \cite{wen2017terngrad} quantize each gradient coordinate; \cite{seide20141,Carlson:2015to,bernstein2018signsgd,bernstein2019iclr} and \cite{karimireddy2019error} replace each coordinate of the gradient with its sign; \cite{lin2017deep, stich2018sparsified} and \cite{wangni2018gradient} use the largest few coordinates; and \cite{konevcny2016bfederated} and \cite{wang2018atomo} use a low-rank approximation.

\atomo\ by \cite{wang2018atomo} is perhaps the closest to our work.
It performs importance sampling of the gradient's singular vectors and is an unbiased compression scheme.
It requires, however, a full Singular Value Decomposition every iteration and is hence computationally impractical.

\paragraph{Commutative compression and addition}
\cite{yu2018gradiveq} stress that commutability of compression with gradient addition enables efficient aggregation with \emph{ring all-reduce}. 
Most compressors, however, lack this property.
Yu et al.\ utilize temporally-consistent correlations between gradients coordinates to compress them linearly.
\powersgd\ has a similar property that we call `linearity'.

\paragraph{Error feedback}
First introduced in \citep{seide20141} and analyzed in \citep{stich2018sparsified} for the convex case, error feedback involves computing the difference between a worker's gradient and the compressed gradient (\ie\ \emph{error}) and adding it back to the next gradient (\emph{feedback}).
\cite{karimireddy2019error} and \cite{stich2019errorfeedback} further develop and generalize the framework of error feedback with improved rates.
In the non-convex setting, \cite{karimireddy2019error} show that error feedback is crucial both for convergence and generalization when using biased compressors (e.g. sign or top-$K$).
In general, biased compression schemes equipped with error feedback tend to out-perform their unbiased counterparts.
The practical algorithm by \cite{lin2017deep} is also as an approximate top-$K$ compressor with error feedback.

\paragraph{Low-rank methods}
Recent works argue that in modern over-parameterized deep networks, the final model learnt has a `low stable rank'~\citep{martin2018implicit, li2017algorithmic}. This can partially explain their impressive generalization properties despite being substantially overparameterized \citep{arora2018stronger}.
Adding explicit spectral regularization has shown to further improve the performance of such models \citep{mazumder2010spectral, yoshida2017spectral}.
Using a low-rank update (as we do) can be viewed as implicitly performing a similar regularization~\citep{gunasekar2018characterizing}.
If the target matrices are known to be exactly low-ranked (instead of just low stable rank), \cite{yurtsever2017sketchy} show that it is sometimes possible to converge to the optima using low rank approximations of the gradients without the need for error feedback.


\section{Method}\label{sec:method}
In data-parallel optimization of machine learning models, a number of $W$ workers share the same model parameters $\xx \in \R^d$.
They iteratively update $\xx$ by computing independent stochastic gradients, aggregating these gradients by averaging\footnote{\cite{bernstein2019iclr} propose \signum\ which aggregates 1-bit gradients by majority voting instead of averaging.}, and updating the model parameters based on this aggregate.

\vspace{-2mm}%

\begin{algorithm}
\caption{Rank-$r$ \powersgd\ compression}\label{alg:powersgd}
\begin{algorithmic}[1]
\State The update vector $\Delta_w$ is treated as a list of tensors corresponding to individual model parameters. Vector-shaped parameters (biases) are aggregated uncompressed. Other parameters are reshaped into matrices. The functions below operate on such matrices independently. For each matrix $M \in \R^{n \times m}$, a corresponding  $Q \in \R^{m\times r}$ is initialized from an \iid\ standard normal distribution.

\Function{compress+aggregate}{update matrix $M \in \R^{n \times m}$, previous $Q \in \R^{m\times r}$}
    \State $P \leftarrow M Q$
    \State $P \leftarrow \textsc{all reduce mean}(P)$
            \Comment Now, $P = \frac{1}{W}(M_1+\ldots+M_W) Q$
    \State $\hat{P} \leftarrow \textsc{orthogonalize}(P)$
            \Comment Orthonormal columns
    \State $Q \leftarrow M^\top \hat{P}$
    \State $Q \leftarrow \textsc{all reduce mean}(Q)$
        \Comment Now, $Q = \frac{1}{W}(M_1+\ldots+M_W)^\top \hat{P}$
    \State \Return the compressed representation $(\hat{P}, Q)$.
    \EndFunction
\Function{decompress}{$\hat P \in \R^{n\times r}$, $Q \in \R^{m \times r}$}
    \State \Return $\hat{P} Q^\top$
\EndFunction
\end{algorithmic}
\end{algorithm}

\vspace{-4mm}%
\paragraph{\powersgd\ compression}
We approximate each layer in the model independently.
The parameters of fully-connected layers (dense matrix multiplication) and their gradients have an inherent matrix structure.
The parameters of convolutional layers can be naturally interpreted as fully-connected layers applied repeatedly over a 2D grid of inputs. Practically, this amounts to flattening input and kernel dimensions in the 4D gradient tensors.
Neural networks also contain bias vectors, but these typically constitute a tiny fraction of the parameter space and can be aggregated uncompressed.

For each parameter's gradient $M \in \R^{n \times m}$, the aim of rank-$r$ matrix approximation is to find matrices $P \in \R^{n \times r}$ and $Q \in \R^{m \times r}$ such that $P Q^\top$ approximates $M$ well.
\powersgd\ uses a single step of subspace iteration---\emph{power} iteration generalized to $r>1$---to compute such an approximation.
This involves performing one right multiplication, one left multiplication, and an orthogonalization. We use the Gram-Schmidt procedure to orthogonalize our matrices since they have very few columns (1--4), and this is the most expensive part of the compression procedure.
Further, we `warm-start' the subspace iteration by reusing the approximation computed at the previous step.
With the inclusion of warm-start, a \emph{single} step of subspace iteration yields a factorization $M\sim P Q^\top$ with the same performance as the best rank-$r$ approximation from an expensive Singular Value Decomposition.

%

\paragraph{Efficient aggregation between workers}
In data-parallel optimization, we want to approximate the \emph{average} of the worker's gradients. Suppose \powersgd\ operates on a list of corresponding gradients $\left[M_1\ldots M_W\right]$ from $W$ workers. Both occurrences of $M$ in the algorithm are a (linear) matrix multiplication followed by a (linear) mean reduction over workers.
This introduces a practical invariance: execution on 1 worker with batch size $B\times W$ is equivalent to execution on $W$ workers with batch size $B$ each. We call this property `linearity'. Refer to Appendix~\ref{subsec:single-multi} for more details.

\begin{wrapfigure}[7]{r}[0pt]{0.44\linewidth}
  \begin{subfigure}[t]{0.2\textwidth}
    \includegraphics[width=\textwidth]{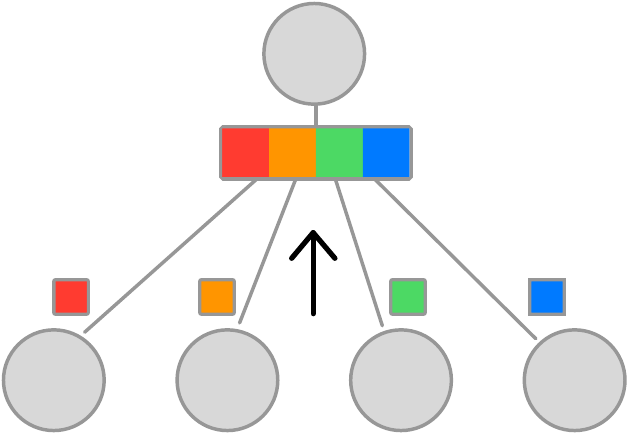}
    \caption{Gather}
  \end{subfigure}
  \hspace{12pt}
  \begin{subfigure}[t]{0.2\textwidth}
    \includegraphics[width=\textwidth]{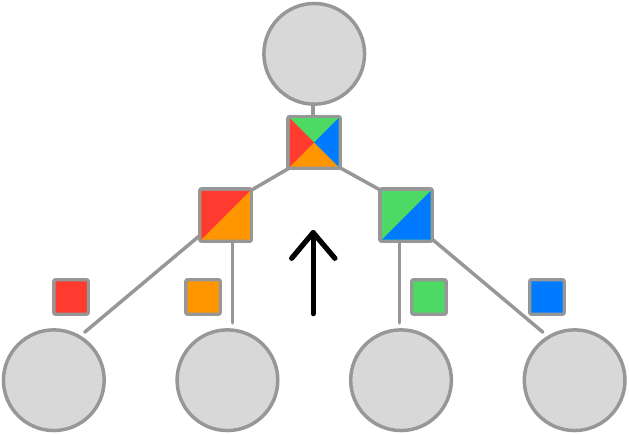}
    \caption{Reduce}
  \end{subfigure}
\end{wrapfigure}

An important benefit of the \powersgd's linearity is that it can be implemented using the \textbf{\allreduce} protocol as opposed to needing a gather operation.
To illustrate the difference, suppose that we want to compute the sum of $W$ matrices $\sum_{i=1}^W M_i$ for $W = 4$.
The \allreduce\ method can use associativity of addition to rewrite the computation as $(M_1 + M_2) + (M_3 + M_4)$.
This enables a divide-and-conquer approach and allows the summation task to be split over multiple workers, as illustrated on the right.
With $W$ workers, both the computation and the communication time scale as $\cO(\log W)$ for {\allreduce}, compared to $\cO(W)$ for {\allgather}. 

In addition to improved scaling, all-reduce communication is preferred over a parameter-server setting because it avoids \emph{double compression}. With a parameter server, both the `clients $\rightarrow$ server' and `server $\rightarrow$ clients' communication have to be compressed~\citep{calas2018expanding,bernstein2019iclr,seide20141}. We avoid this by merging compression and aggregation into one step.

\paragraph{Error-feedback SGD}
Since the \powersgd\ scheme is biased (\ie\ compressing and decompressing a random gradient does not yield the original in expectation), we use error feedback~\citep{seide20141,karimireddy2019error}.
Our version of error feedback (Algorithm~\ref{alg:ef_sgd}) extends the original by introducing post-compression \emph{momentum}.
This simple extension allows us to reuse the same learning rate and hyper-parameters as those tuned for SGD with momentum.

\begin{algorithm}
\caption{Distributed Error-feedback SGD with Momentum}\label{alg:ef_sgd}
\begin{algorithmic}[1]
    \Hyperparameters{learning rate $\gamma$, momentum parameter $\lambda$} 
    \Initialize{model parameters $\xx \in \R^d$, momentum $\mm \leftarrow \0 \in \R^d$, replicated across workers}
    \At{each worker $w=1,\ldots,W$}
    \Initialize{memory $\ee_w \leftarrow \0 \in \R^d$}
    \For{each iterate $t=0,\ldots$}
    \State Compute a stochastic gradient $\gg_w \in \R^d$.
    \State \makebox[10mm][l]{$\Delta_w       $} $\leftarrow \gg_w + \ee_w$
            \Comment{Incorporate error-feedback into update}
    \State \makebox[10mm][l]{$\cC(\Delta_w)$} $\leftarrow \textsc{compress}(\Delta_w)$
    \State \makebox[10mm][l]{$\ee_w          $} $\leftarrow \Delta_w - \textsc{decompress}(\cC(\Delta_w))$
            \Comment{Memorize local errors}
    \State \makebox[10mm][l]{$\cC(\Delta)   $} $\leftarrow \textsc{aggregate}(\cC(\Delta_1), \ldots, \cC(\Delta_W))$
            \Comment{Exchange gradients}
    \State \makebox[10mm][l]{$\Delta^\prime  $} $\leftarrow \textsc{decompress}(\cC(\Delta))$
            \Comment{Reconstruct an update $\in \R^d$}
    \State \makebox[10mm][l]{$\mm            $} $\leftarrow \lambda\mm + \Delta^\prime$
    \State \makebox[10mm][l]{$\xx            $} $\leftarrow \xx - \gamma\,(\Delta^\prime + \mm)$
    \EndFor
    \EndAt
    \end{algorithmic}
\end{algorithm}




\section{Analysis of \powersgd}\label{sec:ablation_study}
In this section, we consider different aspects of \powersgd\ in isolation and hope to empirically understand:
i) the effect of using error feedback,
ii) the effect of `warm-start', and
iii) the trade-off between test accuracy and compression rate with varying approximation rank.

\subsection{Effect of error feedback}
Using error-feedback SGD as a base algorithm for \powersgd\ has two advantages. First, it enables our use of a biased compressor. 
Secondly, EF-SGD improves convergence and obtains better test accuracy~\citep{karimireddy2019error}.

To illustrate the improved test accuracy, we compare \powersgd---a biased compressor with error feedback---against an unbiased low-rank approximation.
To approximate a matrix $M \in \real^{n \times m}$, the unbiased rank-$r$ approximator samples a random matrix $U \in \real^{m \times r}$ such that $\expect[U U^\top] = I_m$ and outputs $(MU,U)$ as the low-rank approximation.
This scheme is unbiased since
\[
	\expect[(MU)U^\top] = M\expect[U U^\top] = MI = M\,.
\]
\powersgd\ is the natural biased counterpart of this unbiased scheme.
Table~\ref{tab:rank_based_ef} demonstrates that our biased approximator with error feedback outperforms the unbiased operator on image classification.
\begin{table}
  \centering
  \begin{minipage}[t]{0.55\textwidth}
    \caption{
      Rank-based compression with and without error feedback. The biased \powersgd\ outperforms an unbiased linear rank-$r$ compressor on test accuracy.
    }
    \label{tab:rank_based_ef}
    \vspace{1mm}
\newcolumntype{R}{>{\raggedleft\arraybackslash}X}
\tablefontsize
\begin{tabularx}{1\linewidth}{Xlr}
    \toprule
    Algorithm 
    & Test accuracy
    & \multicolumn{1}{c}{Data/epoch} \\
    \cmidrule(lr){1-1} \cmidrule(lr){2-2} \cmidrule(lr){3-3}SGD& $94.3\%$ \tikz{
\draw[gray,line width=.3pt] (0,0) -- (1.2,0);
\draw[white, line width=0.01pt] (0,-2pt) -- (0,2pt);
\draw[black,line width=1pt] (1.1506403160095215,0) -- (1.176720371246338,0);
\draw[black,line width=1pt] (1.1506403160095215,-2pt) -- (1.1506403160095215,2pt);
\draw[black,line width=1pt] (1.176720371246338,-2pt) -- (1.176720371246338,2pt);}
    & 1023 MB \\Rank-1 \powersgd& $93.6\%$ \tikz{
\draw[gray,line width=.3pt] (0,0) -- (1.2,0);
\draw[white, line width=0.01pt] (0,-2pt) -- (0,2pt);
\draw[black,line width=1pt] (1.1225603485107423,0) -- (1.1370401573181153,0);
\draw[black,line width=1pt] (1.1225603485107423,-2pt) -- (1.1225603485107423,2pt);
\draw[black,line width=1pt] (1.1370401573181153,-2pt) -- (1.1370401573181153,2pt);}
    & 4 MB \\Rank-2 \powersgd& $94.4\%$ \tikz{
\draw[gray,line width=.3pt] (0,0) -- (1.2,0);
\draw[white, line width=0.01pt] (0,-2pt) -- (0,2pt);
\draw[black,line width=1pt] (1.1656004905700683,0) -- (1.1748803901672367,0);
\draw[black,line width=1pt] (1.1656004905700683,-2pt) -- (1.1656004905700683,2pt);
\draw[black,line width=1pt] (1.1748803901672367,-2pt) -- (1.1748803901672367,2pt);}
    & 8 MB \\Unbiased Rank 1& $71.2\%$ \tikz{
\draw[gray,line width=.3pt] (0,0) -- (1.2,0);
\draw[white, line width=0.01pt] (0,-2pt) -- (0,2pt);
\draw[black,line width=1pt] (0.04216020584106488,0) -- (0.08552026748657227,0);
\draw[black,line width=1pt] (0.04216020584106488,-2pt) -- (0.04216020584106488,2pt);
\draw[black,line width=1pt] (0.08552026748657227,-2pt) -- (0.08552026748657227,2pt);}
    & 3 MB \\Unbiased Rank 2& $75.9\%$ \tikz{
\draw[gray,line width=.3pt] (0,0) -- (1.2,0);
\draw[white, line width=0.01pt] (0,-2pt) -- (0,2pt);
\draw[black,line width=1pt] (0.23728019714355478,0) -- (0.3316802215576173,0);
\draw[black,line width=1pt] (0.23728019714355478,-2pt) -- (0.23728019714355478,2pt);
\draw[black,line width=1pt] (0.3316802215576173,-2pt) -- (0.3316802215576173,2pt);}
    & 4 MB \\
    \bottomrule
\end{tabularx}
  \end{minipage}
  \hspace{12pt}
  \begin{minipage}[t]{0.4\textwidth}
    \caption{Best rank-2 approximation vs.\ \powersgd. Warm-start improves test accuracy, even matching the performance of the best rank-2 approximation.}
    \vspace{11pt}
	\label{tab:warm_start}
\newcolumntype{R}{>{\raggedleft\arraybackslash}X}
\tablefontsize
\begin{tabularx}{1\linewidth}{Xr}
    \toprule
    Algorithm 
    & \multicolumn{1}{l}{Test accuracy}\\
    \cmidrule(lr){1-1} \cmidrule(lr){2-2}
    Best approximation
    & $94.4\%$ \tikz{
\draw[gray,line width=.3pt] (0,0) -- (1.2,0);
\draw[white, line width=0.01pt] (0,-2pt) -- (0,2pt);
\draw[black,line width=1pt] (0.8254600004716395,0) -- (1.1054618141867893,0);
\draw[black,line width=1pt] (0.8254600004716395,-2pt) -- (0.8254600004716395,2pt);
\draw[black,line width=1pt] (1.1054618141867893,-2pt) -- (1.1054618141867893,2pt);} \\
    Warm start {\color{gray}(default)}
    & $94.4\%$ \tikz{
\draw[gray,line width=.3pt] (0,0) -- (1.2,0);
\draw[white, line width=0.01pt] (0,-2pt) -- (0,2pt);
\draw[black,line width=1pt] (0.8545566038651872,0) -- (1.0654634128917435,0);
\draw[black,line width=1pt] (0.8545566038651872,-2pt) -- (0.8545566038651872,2pt);
\draw[black,line width=1pt] (1.0654634128917435,-2pt) -- (1.0654634128917435,2pt);} \\
    Without warm start
    & $94.0\%$ \tikz{
\draw[gray,line width=.3pt] (0,0) -- (1.2,0);
\draw[white, line width=0.01pt] (0,-2pt) -- (0,2pt);
\draw[black,line width=1pt] (0.3472803289240006,0) -- (0.8654597022316693,0);
\draw[black,line width=1pt] (0.3472803289240006,-2pt) -- (0.3472803289240006,2pt);
\draw[black,line width=1pt] (0.8654597022316693,-2pt) -- (0.8654597022316693,2pt);} \\
    \bottomrule
\end{tabularx}
  \end{minipage}
\end{table}

\subsection{Effect of warm-start}
\powersgd\ does not compute the best rank-$r$ approximation of a gradient matrix, but uses a cheaper, low-fidelity approximation based on power iteration.
Comparing the time per batch of \powersgd\ and \atomo\ in Table~\ref{tab:against_others}, we see the importance of avoiding a Singular Value Decomposition.
With gradients shaped as in \powersgd, computing the SVD of a stochastic gradient takes 673ms, the equivalent of computing 6 mini-batch gradients.
In contrast, one full step of rank-2 \powersgd, including communication between 16 workers, takes only 105ms.

Given that we only use a single step of power iteration, the quality of the approximation suffers---compare the test accuracy of `without warm start' and `best approximation' in Table~\ref{tab:warm_start}.
A key feature of \powersgd\ is the \emph{warm start} strategy which reuses previously computed matrix approximations to initialize the power iteration algorithm.
If the matrix on which we perform power iteration remains constant, then this recovers the best rank-$r$ approximation (see Theorem~\ref{thm:subspace-converge} in the Appendix).
We argue that this strategy sometimes makes sense even if the underlying matrices are varying.

Suppose we approximate the sequence of gradient matrices $\{M_t\}$ at timesteps $t$.
At timestep $t$, we leverage the previous factorization $M_{t-1} \approx P_{t-1} Q_{t-1}^\top$.
If $M_t \approx M_{t-1}$ then we would benefit from reusing $P_{t-1}$ and $Q_{t-1}$ as our starting point.
While this is unlikely to be true, if $M_t$ and $M_{t-1}$ are stochastic approximations of the full gradient, we can expect that $\expect[M_t] \approx \expect[M_{t-1}]$ since the function is smooth and we only take small update steps.
The result is akin to Oja's algorithm for \emph{stochastic power iteration}~\citep{oja1982simplified}, and hence could result in an improved approximation quality.
As we show empirically in Table~\ref{tab:warm_start}, this `warm starting' strategy is sufficient to close the gap in test accuracy between \powersgd\ and the much more expensive best rank-$r$ approximation.

\subsection{Effect of varying the rank}
\label{subsec:ablation_rank}
\powersgd\ allows users to choose the rank of its gradient approximations.
The trade-off between approximation quality and compression, decompression and transfer cost is explored in Table~\ref{tab:which_rank}.
In both the image classification and language modeling tasks we explore, the test quality achieved by \powersgd\ grows with increasing rank.
In both cases, it reaches a quality that is as good, or even slightly better than regular SGD.

\begin{table}
  \begin{minipage}[t]{.28\textwidth}\vspace{3mm}
  \caption{
    \powersgd\ with varying rank.
    With sufficient rank, \powersgd\ accelerates training of a \resnet\ and an LSTM by reducing communication, achieving test quality on par with regular SGD in the same number of iterations.
    The time per batch includes the forward/backward pass (constant).
    See Section~\ref{sec:results} for the experimental setup.
  }
  \label{tab:which_rank}%
  \end{minipage}%
  \hspace{.02\textwidth}%
  \begin{minipage}[t]{.7\textwidth}
    \vspace{1.5mm}
    \tablefontsize
    \centering\textbf{Image classification --- \resnet\ on \cifar} \\[1pt]
\newcolumntype{R}{>{\raggedleft\arraybackslash}X}
\tablefontsize
\begin{tabularx}{1\linewidth}{Xlrlrr}
    \toprule
    Algorithm 
    & Test accuracy
    & \multicolumn{2}{l}{Data sent per epoch} 
    & \multicolumn{2}{l}{Time per batch}\\
    \cmidrule(lr){1-1} \cmidrule(lr){2-2} \cmidrule(lr){3-4} \cmidrule(lr){5-6}
    SGD
    & $94.3\%$ \tikz{
\draw[gray,line width=.3pt] (0,0) -- (1.2,0);
\draw[white, line width=0.01pt] (0,-2pt) -- (0,2pt);
\draw[black,line width=1pt] (0.8915019750595093,0) -- (1.0545023202896129,0);
\draw[black,line width=1pt] (0.8915019750595093,-2pt) -- (0.8915019750595093,2pt);
\draw[black,line width=1pt] (1.0545023202896129,-2pt) -- (1.0545023202896129,2pt);}
    & 1023 MB
    & {\color{gray}($1 \times$)}
    & 312 ms 
    & {\color{gray}$+0\%$}\\
    Rank 1
    & $93.6\%$ \tikz{
\draw[gray,line width=.3pt] (0,0) -- (1.2,0);
\draw[white, line width=0.01pt] (0,-2pt) -- (0,2pt);
\draw[black,line width=1pt] (0.7160021781921391,0) -- (0.8065009832382201,0);
\draw[black,line width=1pt] (0.7160021781921391,-2pt) -- (0.7160021781921391,2pt);
\draw[black,line width=1pt] (0.8065009832382201,-2pt) -- (0.8065009832382201,2pt);}
    & 4 MB
    & {\color{gray}($243 \times$)}
    & 229 ms 
    & {\color{gray}$-26\%$}\\
    Rank 2
    & $94.4\%$ \tikz{
\draw[gray,line width=.3pt] (0,0) -- (1.2,0);
\draw[white, line width=0.01pt] (0,-2pt) -- (0,2pt);
\draw[black,line width=1pt] (0.9850030660629268,0) -- (1.0430024385452294,0);
\draw[black,line width=1pt] (0.9850030660629268,-2pt) -- (0.9850030660629268,2pt);
\draw[black,line width=1pt] (1.0430024385452294,-2pt) -- (1.0430024385452294,2pt);}
    & 8 MB
    & {\color{gray}($136 \times$)}
    & 239 ms 
    & {\color{gray}$-23\%$}\\
    Rank 4
    & $94.5\%$ \tikz{
\draw[gray,line width=.3pt] (0,0) -- (1.2,0);
\draw[white, line width=0.01pt] (0,-2pt) -- (0,2pt);
\draw[black,line width=1pt] (1.024501276016235,0) -- (1.0565018177032468,0);
\draw[black,line width=1pt] (1.024501276016235,-2pt) -- (1.024501276016235,2pt);
\draw[black,line width=1pt] (1.0565018177032468,-2pt) -- (1.0565018177032468,2pt);}
    & 14 MB
    & {\color{gray}($72 \times$)}
    & 260 ms 
    & {\color{gray}$-16\%$}\\
    \bottomrule
\end{tabularx}
    \vspace{1mm}\\
    \centering\textbf{Language modeling --- LSTM on \wikitext} \\[1pt]
\newcolumntype{R}{>{\raggedleft\arraybackslash}X}
\tablefontsize
\begin{tabularx}{1\linewidth}{Xrrlrr}
    \toprule
    Algorithm 
    & \multicolumn{1}{l}{Test perplexity} 
    & \multicolumn{2}{l}{Data sent per epoch} 
    & \multicolumn{2}{l}{Time per batch}\\
    \cmidrule(lr){1-1} \cmidrule(lr){2-2} \cmidrule(lr){3-4} \cmidrule(lr){5-6}
    SGD
    & 91 \tikz{
\draw[gray,line width=.3pt] (0,0) -- (1.2,0);
\draw[white, line width=0.01pt] (0,-2pt) -- (0,2pt);
\draw[black,line width=1pt] (0.25200451660156203,0) -- (0.2751595458984377,0);
\draw[black,line width=1pt] (0.25200451660156203,-2pt) -- (0.25200451660156203,2pt);
\draw[black,line width=1pt] (0.2751595458984377,-2pt) -- (0.2751595458984377,2pt);}
    & 7730 MB
    & {\color{gray}($1 \times$)}
    & 300 ms 
    & {\color{gray}$+0\%$}\\
    Rank 1
    & 102 \tikz{
\draw[gray,line width=.3pt] (0,0) -- (1.2,0);
\draw[white, line width=0.01pt] (0,-2pt) -- (0,2pt);
\draw[black,line width=1pt] (1.0398884277343745,0) -- (1.1214045410156246,0);
\draw[black,line width=1pt] (1.0398884277343745,-2pt) -- (1.0398884277343745,2pt);
\draw[black,line width=1pt] (1.1214045410156246,-2pt) -- (1.1214045410156246,2pt);}
    & 25 MB
    & {\color{gray}($310 \times$)}
    & 131 ms 
    & {\color{gray}$-56\%$}\\
    Rank 2
    & 93 \tikz{
\draw[gray,line width=.3pt] (0,0) -- (1.2,0);
\draw[white, line width=0.01pt] (0,-2pt) -- (0,2pt);
\draw[black,line width=1pt] (0.4124176025390625,0) -- (0.46680749511718767,0);
\draw[black,line width=1pt] (0.4124176025390625,-2pt) -- (0.4124176025390625,2pt);
\draw[black,line width=1pt] (0.46680749511718767,-2pt) -- (0.46680749511718767,2pt);}
    & 38 MB
    & {\color{gray}($203 \times$)}
    & 141 ms 
    & {\color{gray}$-53\%$}\\
    Rank 4
    & 91 \tikz{
\draw[gray,line width=.3pt] (0,0) -- (1.2,0);
\draw[white, line width=0.01pt] (0,-2pt) -- (0,2pt);
\draw[black,line width=1pt] (0.21946130371093772,0) -- (0.2438134765625,0);
\draw[black,line width=1pt] (0.21946130371093772,-2pt) -- (0.21946130371093772,2pt);
\draw[black,line width=1pt] (0.2438134765625,-2pt) -- (0.2438134765625,2pt);}
    & 64 MB
    & {\color{gray}($120 \times$)}
    & 134 ms 
    & {\color{gray}$-55\%$}\\
    \bottomrule
\end{tabularx}
  \end{minipage}
\end{table}

\section{Results} \label{sec:results}

\begin{wraptable}[15]{r}[0pt]{0.45\linewidth}%
\vspace{-20pt}
\scriptsize%
\centering Default experimental setting\\
\vspace{1mm}
\label{tab:defaults}%
\begin{tabularx}{\linewidth}{lX}
    \toprule
    Dataset & \cifar \\
    Architecture & \resnet \\
    \midrule
    Number of workers & 16 \\
    Backend & \nccl\ (fastest in \pytorch) \\
    Batch size & $128 \times \text{number of workers}$ \\
    \midrule
    Momentum & 0.9 \\
    Learning rate & Tuned for 16 workers --- $0.1 \times 16$ for SGD.
     Scaled linearly by the number of workers \\
    LR decay & $/10$ at epoch 150 and 250 \\
    LR warmup & Linearly within 5 epochs, starting from the single-worker LR \\
    \# Epochs & 300 \\
    Weight decay & $10^{-4}$,\\
                  &$0$ for BatchNorm parameters \\
    \midrule
    Repetitions & 3, with varying seeds \\
    Error bars & min --- max \\
    \bottomrule
\end{tabularx}
\end{wraptable}

This section demonstrates the practicality of \powersgd\ for distributed optimization of deep neural networks. We show that the compression scheme of \powersgd\ i) is fast and matches test performance of SGD, ii) scales well with increasing workers even with a sub-optimal communication backend, and iii) significantly reduces training time for larger models.

Most of the analysis is performed on \cifar, in the setting described in the table on the right.
We verify the generality of \powersgd\ by an additional evaluation of an LSTM for language modeling on \wikitext.
We use 16 GPUs on 8 machines, connected through a fast (10Gbit/s) network. To obtain meaningful timings, we have aimed to optimize all compared optimizers to a similar level. We provide a list of our performance optimizations in Appendix~\ref{sec:optimizations}.
Throughout these results, we tune the learning rate for full-precision SGD, and use the \emph{same} parameters for \powersgd\ and other compression algorithms that use error feedback with momentum.
Learning rates for the compared-to \atomo~\citep{wang2018atomo} and \signum~\citep{bernstein2019iclr} were separately tuned \cf~Appendix~{\ref{sec:lr_tuning}}.

\subsection{Comparison with other compressors}
\begin{table}
  \centering
  \begin{minipage}{\textwidth}
    \caption{
      Comparing different compression operators for Error-feedback SGD in a unified setting; running 300 epochs of Error-feedback SGD with Momentum (Algorithm~\ref{alg:ef_sgd}) with a learning rate tuned for full-precision SGD on 16 GPUs for \cifar. Note that the variations of \powersgd\ with ranks~2 and 7 strike the best balance between the achieved test accuracy and time per batch (total time for forward, backward, compression, decompression, and gradient aggregation).
    }
    \label{tab:ef_compressors}
    \vspace{1mm}
\newcolumntype{R}{>{\raggedleft\arraybackslash}X}
\tablefontsize
\begin{tabularx}{1\linewidth}{lXlrcl}
    \toprule
    \multicolumn{2}{c}{}
    & \multicolumn{1}{c}{Test accuracy}
    & \multicolumn{1}{r}{Sent/epoch} 
    & \multicolumn{1}{c}{All-reduce} 
    & \multicolumn{1}{c}{Time/batch}\\
    \cmidrule(lr){3-3} \cmidrule(lr){4-4} \cmidrule(lr){5-5}  \cmidrule(lr){6-6}
    \multicolumn{2}{l}{No compression}
    & $94.3\%$ \tikz{
\draw[gray,line width=.3pt] (0,0) -- (1.2,0);
\draw[white, line width=0.01pt] (0,-2pt) -- (0,2pt);
\draw[black,line width=1pt] (0.7253876539377059,0) -- (0.9761574158301733,0);
\draw[black,line width=1pt] (0.7253876539377059,-2pt) -- (0.7253876539377059,2pt);
\draw[black,line width=1pt] (0.9761574158301733,-2pt) -- (0.9761574158301733,2pt);}
    & 1023 MB
    & \cmark
    & 312 ms
    ~\tikz{\draw[black,line width=.5pt] (0,2pt) -- (0, -2pt); \draw[black,line width=1.5pt] (0,0) -- (0.6978106464907947,0);  \draw[gray,line width=.3pt] (0.6940298507462686,2pt) -- (0.6940298507462686, -2pt);} \\\cmidrule(lr){1-2}
    Medium
    & \textbf{\hspace{0pt}Rank 7\hspace{0pt}}
    & $94.6\%$ \tikz{
\draw[gray,line width=.3pt] (0,0) -- (1.2,0);
\draw[white, line width=0.01pt] (0,-2pt) -- (0,2pt);
\draw[black,line width=1pt] (0.9053872621976413,0) -- (1.0807729207552423,0);
\draw[black,line width=1pt] (0.9053872621976413,-2pt) -- (0.9053872621976413,2pt);
\draw[black,line width=1pt] (1.0807729207552423,-2pt) -- (1.0807729207552423,2pt);}
    & 24 MB
    & \cmark
    & 285 ms
    ~\tikz{\draw[black,line width=.5pt] (0,2pt) -- (0, -2pt); \draw[black,line width=1.5pt] (0,0) -- (0.6373218116872993,0);  \draw[gray,line width=.3pt] (0.6940298507462686,2pt) -- (0.6940298507462686, -2pt);} \\
    
    & Random Block
    & $93.3\%$ \tikz{
\draw[gray,line width=.3pt] (0,0) -- (1.2,0);
\draw[white, line width=0.01pt] (0,-2pt) -- (0,2pt);
\draw[black,line width=1pt] (0.4169275650611295,0) -- (0.4476945656996512,0);
\draw[black,line width=1pt] (0.4169275650611295,-2pt) -- (0.4169275650611295,2pt);
\draw[black,line width=1pt] (0.4476945656996512,-2pt) -- (0.4476945656996512,2pt);}
    & 24 MB
    & \cmark
    & 243 ms
    ~\tikz{\draw[black,line width=.5pt] (0,2pt) -- (0, -2pt); \draw[black,line width=1.5pt] (0,0) -- (0.5440989983628195,0);  \draw[gray,line width=.3pt] (0.6940298507462686,2pt) -- (0.6940298507462686, -2pt);} \\
    
    & Random K
    & $94.0\%$ \tikz{
\draw[gray,line width=.3pt] (0,0) -- (1.2,0);
\draw[white, line width=0.01pt] (0,-2pt) -- (0,2pt);
\draw[black,line width=1pt] (0.6776938511775082,0) -- (0.7723100808950556,0);
\draw[black,line width=1pt] (0.6776938511775082,-2pt) -- (0.6776938511775082,2pt);
\draw[black,line width=1pt] (0.7723100808950556,-2pt) -- (0.7723100808950556,2pt);}
    & 24 MB
    & \cmark
    & 540 ms
    ~\tikz{\draw[black,line width=.5pt] (0,2pt) -- (0, -2pt); \draw[black,line width=1.5pt] (0,0) -- (1.2091594474650624,0);  \draw[gray,line width=.3pt] (0.6940298507462686,2pt) -- (0.6940298507462686, -2pt);} \\
    
    & Sign+Norm
    & $93.9\%$ \tikz{
\draw[gray,line width=.3pt] (0,0) -- (1.2,0);
\draw[white, line width=0.01pt] (0,-2pt) -- (0,2pt);
\draw[black,line width=1pt] (0.5900034977839533,0) -- (0.7707728312565747,0);
\draw[black,line width=1pt] (0.5900034977839533,-2pt) -- (0.5900034977839533,2pt);
\draw[black,line width=1pt] (0.7707728312565747,-2pt) -- (0.7707728312565747,2pt);}
    & 32 MB
    & {\color{lightgray}\xmark}
    & 429 ms
    ~\tikz{\draw[black,line width=.5pt] (0,2pt) -- (0, -2pt); \draw[black,line width=1.5pt] (0,0) -- (0.959362078966268,0);  \draw[gray,line width=.3pt] (0.6940298507462686,2pt) -- (0.6940298507462686, -2pt);} \\
    
    & Top K
    & $94.4\%$ \tikz{
\draw[gray,line width=.3pt] (0,0) -- (1.2,0);
\draw[white, line width=0.01pt] (0,-2pt) -- (0,2pt);
\draw[black,line width=1pt] (0.8869248610276462,0) -- (0.9961576168353733,0);
\draw[black,line width=1pt] (0.8869248610276462,-2pt) -- (0.8869248610276462,2pt);
\draw[black,line width=1pt] (0.9961576168353733,-2pt) -- (0.9961576168353733,2pt);}
    & 32 MB
    & {\color{lightgray}\xmark}
    & 444 ms
    ~\tikz{\draw[black,line width=.5pt] (0,2pt) -- (0, -2pt); \draw[black,line width=1.5pt] (0,0) -- (0.9944611463817571,0);  \draw[gray,line width=.3pt] (0.6940298507462686,2pt) -- (0.6940298507462686, -2pt);} \\\cmidrule(lr){1-2}
    High
    & \textbf{\hspace{0pt}Rank 2\hspace{0pt}}
    & $94.4\%$ \tikz{
\draw[gray,line width=.3pt] (0,0) -- (1.2,0);
\draw[white, line width=0.01pt] (0,-2pt) -- (0,2pt);
\draw[black,line width=1pt] (0.8692354862506562,0) -- (0.9584652900695834,0);
\draw[black,line width=1pt] (0.8692354862506562,-2pt) -- (0.8692354862506562,2pt);
\draw[black,line width=1pt] (0.9584652900695834,-2pt) -- (0.9584652900695834,2pt);}
    & 8 MB
    & \cmark
    & 239 ms
    ~\tikz{\draw[black,line width=.5pt] (0,2pt) -- (0, -2pt); \draw[black,line width=1.5pt] (0,0) -- (0.5351457513464453,0);  \draw[gray,line width=.3pt] (0.6940298507462686,2pt) -- (0.6940298507462686, -2pt);} \\
    
    & Random Block
    & $87.8\%$ \tikz{
\draw[gray,line width=.3pt] (0,0) -- (1.2,0);
\draw[white, line width=0.01pt] (0,-2pt) -- (0,2pt);}
    & 8 MB
    & \cmark
    & 240 ms
    ~\tikz{\draw[black,line width=.5pt] (0,2pt) -- (0, -2pt); \draw[black,line width=1.5pt] (0,0) -- (0.538081557695938,0);  \draw[gray,line width=.3pt] (0.6940298507462686,2pt) -- (0.6940298507462686, -2pt);} \\
    
    & Random K
    & $92.6\%$ \tikz{
\draw[gray,line width=.3pt] (0,0) -- (1.2,0);
\draw[white, line width=0.01pt] (0,-2pt) -- (0,2pt);
\draw[black,line width=1pt] (0.08384837370652246,0) -- (0.13384970151461137,0);
\draw[black,line width=1pt] (0.08384837370652246,-2pt) -- (0.08384837370652246,2pt);
\draw[black,line width=1pt] (0.13384970151461137,-2pt) -- (0.13384970151461137,2pt);}
    & 8 MB
    & \cmark
    & 534 ms
    ~\tikz{\draw[black,line width=.5pt] (0,2pt) -- (0, -2pt); \draw[black,line width=1.5pt] (0,0) -- (1.1947041144766877,0);  \draw[gray,line width=.3pt] (0.6940298507462686,2pt) -- (0.6940298507462686, -2pt);} \\
    
    & Top K
    & $93.6\%$ \tikz{
\draw[gray,line width=.3pt] (0,0) -- (1.2,0);
\draw[white, line width=0.01pt] (0,-2pt) -- (0,2pt);
\draw[black,line width=1pt] (0.5007720433748692,0) -- (0.6400020746084372,0);
\draw[black,line width=1pt] (0.5007720433748692,-2pt) -- (0.5007720433748692,2pt);
\draw[black,line width=1pt] (0.6400020746084372,-2pt) -- (0.6400020746084372,2pt);}
    & 8 MB
    & {\color{lightgray}\xmark}
    & 411 ms
    ~\tikz{\draw[black,line width=.5pt] (0,2pt) -- (0, -2pt); \draw[black,line width=1.5pt] (0,0) -- (0.9204505103738796,0);  \draw[gray,line width=.3pt] (0.6940298507462686,2pt) -- (0.6940298507462686, -2pt);} \\
    \bottomrule
\end{tabularx}
  \end{minipage}
\end{table}

Error feedback in compressed optimization enables the use of a multitude of compression schemes, including biased ones.
The potential compression operators illustrated in Figure~\ref{fig:compression_methods} are compared in Table~\ref{tab:ef_compressors}.
We evaluate compressors based on the test accuracy achieved and the total time taken to process one mini-batch. The former is a holistic measure of the accuracy of the compression operator, and the latter is the net time required for a forward pass, backward pass, gradient compression and decompression and gradient communication.
We study two compression regimes---medium and high.

At around $32\times$ compression, achieved by sign-based methods, all compression schemes (other than Random Block) achieve test accuracy close to full-precision SGD.
This implies that all schemes in this regime (other than Random Block) obtain a good-enough compression quality.
At high compression ($128\times$), \powersgd\ particularly stands out as the only method to achieve the target test accuracy.

In both the medium and high compression settings, the only schemes to be faster than full-precision SGD are \powersgd\ and Random Block.
Note that both are simple linear schemes and hence support \allreduce.
While Random $K$ also supports \allreduce, the overhead for random memory access during both the compression and decompression stages is substantial, making it slower overall than SGD.
Thus, on modern GPU-enabled infrastructure, \powersgd, which relies on matrix multiplication, is faster and much more accurate than the other compression schemes.

\subsection{Scalability of \powersgd}
Here we investigate how \powersgd\ scales with an increasing number of workers, shedding light on what we can expect if we use a significantly larger number of workers. Additionally, we investigate how these results depend on the choice of communication backend.
We benchmark \powersgd\ against SGD and \signum\ (signSGD with majority vote) from \cite{bernstein2019iclr} which we believe is the current state-of-the-art for distributed algorithms.

Table~\ref{tab:timing_breakdown} provides a detailed breakdown of the time spent for each mini-batch (\ie\ one step) into the forward pass, backward pass, gradient exchange (communication), and compression/decompression.
The time spent in the forward and backward pass is constant across all algorithms and numbers of workers.
Since both SGD and \powersgd\ use \allreduce, the gradient communication time (solid green in Table~\ref{tab:timing_breakdown}) scales gracefully with increasing number of workers.
\signum---which uses \allgather\ instead of \allreduce---has a steeper increase.
It has comparable time to \powersgd\ for 4 workers but becomes more expensive for 16 workers.

There is another, more subtle, consequence of \allreduce\ vs. \allgather\ on the decoding times.
In \allreduce, the \emph{aggregation} step and the \emph{communication} step happen simultaneously.
Each worker receives a pre-aggregated gradient, making the cost of decompression independent of the number of workers.
On the other hand, in \allgather, a worker receives $W$ compressed gradients that need to be individually decompressed and aggregated (either using majority vote or averaging).
The time for decompression with \allgather\ therefore scales linearly with number of workers. This shows when comparing the hatcheted regions in Table~\ref{tab:timing_breakdown}.
This observation speaks to the importance of the reduce operation for scalability.

\begin{table}
  \centering
  \begin{minipage}{\textwidth}
    \DeclareRobustCommand\dotOne{\tikz{\fill[fill=color1] (0,0) rectangle (5pt,5pt);}~}
    \DeclareRobustCommand\dotTwo{\tikz{\fill[fill=color2] (0,0) rectangle (5pt,5pt);}~}
    \DeclareRobustCommand\dotThree{\tikz{\fill[fill=color3] (0,0) rectangle (5pt,5pt);}~}
    \DeclareRobustCommand\dotThreeStriped{\tikz{\fill[fill=color3] (0,0) rectangle (5pt,5pt);\fill[pattern=north east lines, pattern color=black!70!color3] (0,0) rectangle (5pt,5pt);}~}
    \caption{
      Breakdown of time spent (in seconds) in one iteration of \resnet\ training. Because \powersgd\ (Rank 2) uses \allreduce, time spent encoding/decoding gradients is constant.\\
      \dotOne Forward pass,
      \dotTwo Backward pass,
      \dotThree Gradient exchange,
      \dotThreeStriped Encoding and decoding.
    }
    \label{tab:timing_breakdown}
    \vspace{1mm}
\newcolumntype{R}{>{\raggedleft\arraybackslash}X}
\tablefontsize
\begin{tabularx}{1\linewidth}{Xllll}
    \toprule& 2 workers& 4 workers& 8 workers& 16 workers \\
    Rank 2
        & \tikz{
        \fill[fill=color1] (0.0,0) rectangle (0.2249595659817883,0.2);
        \fill[fill=color2] (0.2249595659817883,0) rectangle (0.7910210041398044,0.2);
        \fill[fill=color3] (0.7910210041398044,0) rectangle (1.1728886640255558,0.2);
        \fill[pattern=north east lines, pattern color=black!70!color3] (0.7910210041398044,0) rectangle (0.947058952507254,0.2);
        
    }
        & \tikz{
        \fill[fill=color1] (0.0,0) rectangle (0.21900447692885397,0.2);
        \fill[fill=color2] (0.21900447692885397,0) rectangle (0.773667034540677,0.2);
        \fill[fill=color3] (0.773667034540677,0) rectangle (1.365647536090725,0.2);
        \fill[pattern=north east lines, pattern color=black!70!color3] (0.773667034540677,0) rectangle (0.9419634941731154,0.2);
        
    }
        & \tikz{
        \fill[fill=color1] (0.0,0) rectangle (0.21336344078817723,0.2);
        \fill[fill=color2] (0.21336344078817723,0) rectangle (0.7583041787256387,0.2);
        \fill[fill=color3] (0.7583041787256387,0) rectangle (1.5293636963673645,0.2);
        \fill[pattern=north east lines, pattern color=black!70!color3] (0.7583041787256387,0) rectangle (0.9345197892473837,0.2);
        
    }
        & \tikz{
        \fill[fill=color1] (0.0,0) rectangle (0.20580533351550073,0.2);
        \fill[fill=color2] (0.20580533351550073,0) rectangle (0.7369501242950074,0.2);
        \fill[fill=color3] (0.7369501242950074,0) rectangle (1.624642208689395,0.2);
        \fill[pattern=north east lines, pattern color=black!70!color3] (0.7369501242950074,0) rectangle (0.8983912316145143,0.2);
        
    } \\
    SGD
        & \tikz{
        \fill[fill=color1] (0.0,0) rectangle (0.23071972697427437,0.2);
        \fill[fill=color2] (0.23071972697427437,0) rectangle (0.8032558821637603,0.2);
        \fill[fill=color3] (0.8032558821637603,0) rectangle (1.0454488342867734,0.2);
        \fill[pattern=north east lines, pattern color=black!70!color3] (0.8032558821637603,0) rectangle (0.8599920283399165,0.2);
        
    }
        & \tikz{
        \fill[fill=color1] (0.0,0) rectangle (0.2060319422247217,0.2);
        \fill[fill=color2] (0.2060319422247217,0) rectangle (0.7502982244077666,0.2);
        \fill[fill=color3] (0.7502982244077666,0) rectangle (1.9890841804051702,0.2);
        \fill[pattern=north east lines, pattern color=black!70!color3] (0.7502982244077666,0) rectangle (0.7995346109557786,0.2);
        
    }
        & \tikz{
        \fill[fill=color1] (0.0,0) rectangle (0.20481991340750927,0.2);
        \fill[fill=color2] (0.20481991340750927,0) rectangle (0.7410311987361599,0.2);
        \fill[fill=color3] (0.7410311987361599,0) rectangle (2.0945349435845944,0.2);
        \fill[pattern=north east lines, pattern color=black!70!color3] (0.7410311987361599,0) rectangle (0.7849721125609067,0.2);
        
    }
        & \tikz{
        \fill[fill=color1] (0.0,0) rectangle (0.21039480184613568,0.2);
        \fill[fill=color2] (0.21039480184613568,0) rectangle (0.748235845271718,0.2);
        \fill[fill=color3] (0.748235845271718,0) rectangle (2.168986088949525,0.2);
        \fill[pattern=north east lines, pattern color=black!70!color3] (0.748235845271718,0) rectangle (0.7858707095305848,0.2);
        
    } \\
    Signum
        & \tikz{
        \fill[fill=color1] (0.0,0) rectangle (0.22535667809086316,0.2);
        \fill[fill=color2] (0.22535667809086316,0) rectangle (0.7836748221832777,0.2);
        \fill[fill=color3] (0.7836748221832777,0) rectangle (1.1400115174065908,0.2);
        \fill[pattern=north east lines, pattern color=black!70!color3] (0.7836748221832777,0) rectangle (0.9504986982814408,0.2);
        
    }
        & \tikz{
        \fill[fill=color1] (0.0,0) rectangle (0.22110330194166833,0.2);
        \fill[fill=color2] (0.22110330194166833,0) rectangle (0.7705264726486577,0.2);
        \fill[fill=color3] (0.7705264726486577,0) rectangle (1.3917152872524334,0.2);
        \fill[pattern=north east lines, pattern color=black!70!color3] (0.7705264726486577,0) rectangle (1.0026410539164732,0.2);
        
    }
        & \tikz{
        \fill[fill=color1] (0.0,0) rectangle (0.2063194285062241,0.2);
        \fill[fill=color2] (0.2063194285062241,0) rectangle (0.7369489746109272,0.2);
        \fill[fill=color3] (0.7369489746109272,0) rectangle (1.7532480094115097,0.2);
        \fill[pattern=north east lines, pattern color=black!70!color3] (0.7369489746109272,0) rectangle (1.0742805302418015,0.2);
        
    }
        & \tikz{
        \fill[fill=color1] (0.0,0) rectangle (0.2088017085580676,0.2);
        \fill[fill=color2] (0.2088017085580676,0) rectangle (0.7430495493418104,0.2);
        \fill[fill=color3] (0.7430495493418104,0) rectangle (2.1383896771590947,0.2);
        \fill[pattern=north east lines, pattern color=black!70!color3] (0.7430495493418104,0) rectangle (1.2441275214423346,0.2);
        
    } \\ & \tikz{
        \draw[black] (0,0) -- (2.2,0);
        \draw[black] (0.0,-2pt) -- (0.0,2pt);
        \draw[black] (0.7308970099667775,-2pt) -- (0.7308970099667775,2pt)node[anchor=north] {\tiny$0.1$};
        \draw[black] (1.461794019933555,-2pt) -- (1.461794019933555,2pt)node[anchor=north] {\tiny$0.2$};
        \draw[black] (2.192691029900333,-2pt) -- (2.192691029900333,2pt)node[anchor=north] {\tiny$0.3$};
    } & \tikz{
        \draw[black] (0,0) -- (2.2,0);
        \draw[black] (0.0,-2pt) -- (0.0,2pt);
        \draw[black] (0.7308970099667775,-2pt) -- (0.7308970099667775,2pt)node[anchor=north] {\tiny$0.1$};
        \draw[black] (1.461794019933555,-2pt) -- (1.461794019933555,2pt)node[anchor=north] {\tiny$0.2$};
        \draw[black] (2.192691029900333,-2pt) -- (2.192691029900333,2pt)node[anchor=north] {\tiny$0.3$};
    } & \tikz{
        \draw[black] (0,0) -- (2.2,0);
        \draw[black] (0.0,-2pt) -- (0.0,2pt);
        \draw[black] (0.7308970099667775,-2pt) -- (0.7308970099667775,2pt)node[anchor=north] {\tiny$0.1$};
        \draw[black] (1.461794019933555,-2pt) -- (1.461794019933555,2pt)node[anchor=north] {\tiny$0.2$};
        \draw[black] (2.192691029900333,-2pt) -- (2.192691029900333,2pt)node[anchor=north] {\tiny$0.3$};
    } & \tikz{
        \draw[black] (0,0) -- (2.2,0);
        \draw[black] (0.0,-2pt) -- (0.0,2pt);
        \draw[black] (0.7308970099667775,-2pt) -- (0.7308970099667775,2pt)node[anchor=north] {\tiny$0.1$};
        \draw[black] (1.461794019933555,-2pt) -- (1.461794019933555,2pt)node[anchor=north] {\tiny$0.2$};
        \draw[black] (2.192691029900333,-2pt) -- (2.192691029900333,2pt)node[anchor=north] {\tiny$0.3$};
    }\vspace{-1mm} \\
    \bottomrule
\end{tabularx}
  \end{minipage}
\end{table}

We next study two different backends---the more optimized \nccl\, and the slower \gloo.
All three methods scale reasonably well with the optimized \nccl\ backend, although \signum\ has a slope less than 1 in the log-log plot, indicating sub-linear scaling. On the slower \gloo\ backend, \powersgd\ is notably the only method that retains excellent scaling due to its high compression rate.

\begin{figure}
  \centering
  \includegraphics[width=.8\linewidth]{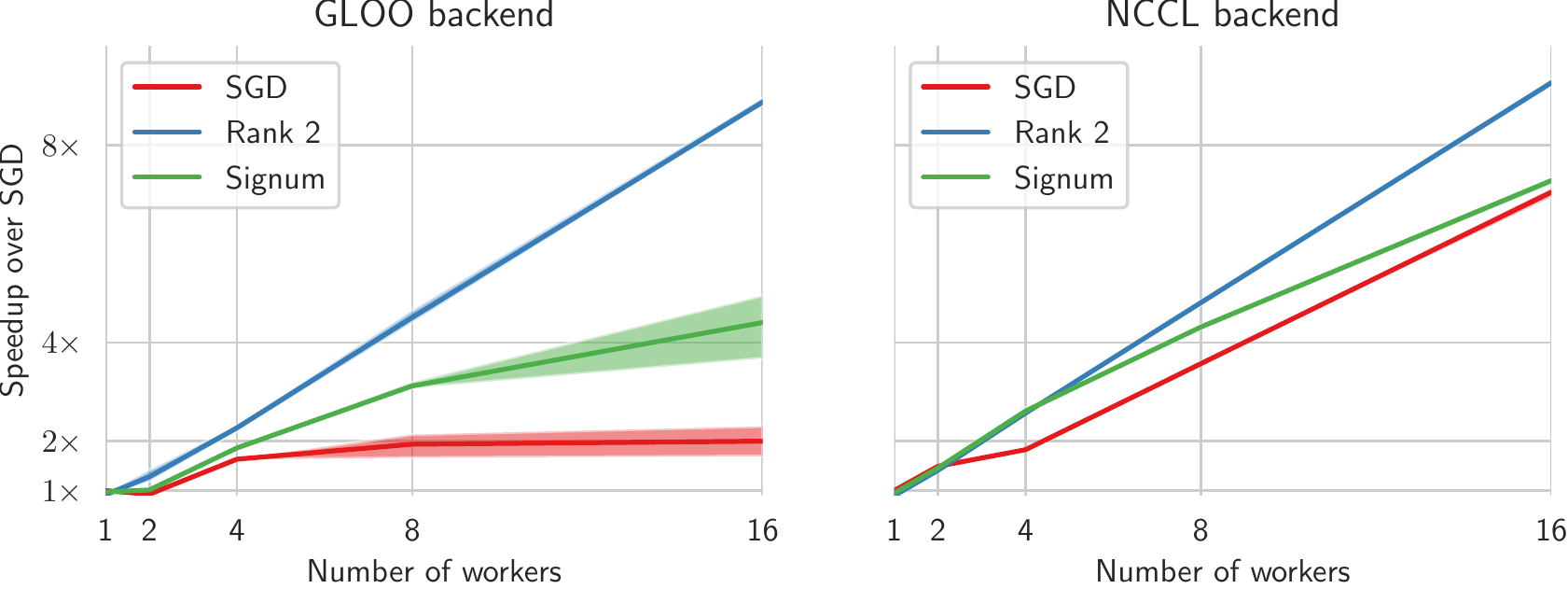}
  \caption{
    Scaling of \powersgd\ on \cifar\ compared to full-precision SGD and \signum~\citep{bernstein2019iclr} on two communication backends.
    The batch size increases linearly with the number of workers.
    We compare training time for one epoch to 1-worker SGD.
    Note that the faster \nccl\ backend used throughout benefits the baselines more than our method.
  }
  \label{fig:scaling}
\end{figure}

\subsection{Other tasks and methods}
In Table~\ref{tab:against_others}, we compare \powersgd\ against the state-of-the-art compressed optimization algorithms \signum\ and \atomo.
The cost of performing a full SVD at each step renders \atomo\ impractical in a high-performance setting, especially considering that it fails to match the test accuracies of the other methods.
\signum\ performs much better, proving a minor speedup over SGD.
\powersgd\ is the fastest and most accurate of the compared methods.

The advantage of \powersgd\ truly shows when using really large models, \ie\ where the communication actually becomes a bottleneck.
To verify this, we run \signum, full-precision SGD, and \powersgd\ to train an LSTM on a language modeling task which has a substantially larger model size than \resnet\ (see Appendix~\ref{sec:network-parameters}).
To match the test score of full-precision SGD, we needed to use a rank-4 approximation (see Section \ref{subsec:ablation_rank}).
\powersgd\ reduces communication by $90\%$ and the overall running time by $55\%$, while \signum\ becomes slower than full-precision SGD and also obtains a worse test score.

Convergence curves on test accuracy corresponding to Tables~\ref{tab:which_rank}, \ref{tab:against_others} and \ref{tab:against_others_lm} are provided in Appendix~\ref{app:convergence_curves}.
In those figures, you can read our improvements in time-to-accuracy for any target accuracy.
We also provide a case study on using PowerSGD for a novel task (language modeling with transformers on \wikitext) and more workers (32) on the public cloud in Appendix~\ref{app:fairseq}.

\begin{table}
  \begin{minipage}[c]{0.33\textwidth}
    \caption{
      Results on \cifar.
      Contrary to rank-2 \atomo~\citep{wang2018atomo} and \signum~\citep{bernstein2019iclr}, \powersgd\ achieves the same test accuracy as full-precision SGD within the default epoch budget.
    }
    \label{tab:against_others}
  \end{minipage}%
  \hspace{.02\textwidth}%
  \begin{minipage}[c]{0.65\textwidth}
    \vspace{1mm}
\newcolumntype{R}{>{\raggedleft\arraybackslash}X}
\tablefontsize
\begin{tabularx}{1\linewidth}{Xrrrr}
    \toprule
    Algorithm 
    & \multicolumn{1}{l}{Test accuracy}
    & \multicolumn{1}{l}{Data/epoch} 
    & \multicolumn{2}{l}{Time per batch}\\
    \cmidrule(lr){1-1} \cmidrule(lr){2-2} \cmidrule(lr){3-3} \cmidrule(lr){4-5}SGD& $94.3\%$ \tikz{
\draw[gray,line width=.3pt] (0,0) -- (1.2,0);
\draw[white, line width=0.01pt] (0,-2pt) -- (0,2pt);
\draw[black,line width=1pt] (0.8721773479295813,0) -- (1.1556562091993272,0);
\draw[black,line width=1pt] (0.8721773479295813,-2pt) -- (0.8721773479295813,2pt);
\draw[black,line width=1pt] (1.1556562091993272,-2pt) -- (1.1556562091993272,2pt);}
    & 1023 MB
    & 312 ms 
    & {\color{gray}$+0\%$}\\Atomo& $92.6\%$ \tikz{
\draw[gray,line width=.3pt] (0,0) -- (1.2,0);
\draw[white, line width=0.01pt] (0,-2pt) -- (0,2pt);
\draw[black,line width=1pt] (0.08695840420930093,0) -- (0.2573932730633256,0);
\draw[black,line width=1pt] (0.08695840420930093,-2pt) -- (0.08695840420930093,2pt);
\draw[black,line width=1pt] (0.2573932730633256,-2pt) -- (0.2573932730633256,2pt);}
    & 113 MB
    & 948 ms 
    & {\color{gray}$+204\%$}\\Signum& $93.6\%$ \tikz{
\draw[gray,line width=.3pt] (0,0) -- (1.2,0);
\draw[white, line width=0.01pt] (0,-2pt) -- (0,2pt);
\draw[black,line width=1pt] (0.6304377016813861,0) -- (0.7365250545999308,0);
\draw[black,line width=1pt] (0.6304377016813861,-2pt) -- (0.6304377016813861,2pt);
\draw[black,line width=1pt] (0.7365250545999308,-2pt) -- (0.7365250545999308,2pt);}
    & 32 MB
    & 301 ms 
    & {\color{gray}$-3\%$}\\\textbf{\hspace{0pt}Rank 2\hspace{0pt}}& $94.4\%$ \tikz{
\draw[gray,line width=.3pt] (0,0) -- (1.2,0);
\draw[white, line width=0.01pt] (0,-2pt) -- (0,2pt);
\draw[black,line width=1pt] (1.0347879409790035,0) -- (1.135656414861269,0);
\draw[black,line width=1pt] (1.0347879409790035,-2pt) -- (1.0347879409790035,2pt);
\draw[black,line width=1pt] (1.135656414861269,-2pt) -- (1.135656414861269,2pt);}
    & 8 MB
    & 239 ms 
    & {\color{gray}$-23\%$}\\
    \bottomrule
\end{tabularx}
  \end{minipage}
\end{table}
\begin{table}[!t]
  \vspace{-1.2\baselineskip}
  \begin{minipage}[c]{0.33\textwidth}
    \caption{
      In \textbf{language modeling}, rank-4 \powersgd\ achieves the target test accuracy and provides a significant speedup over SGD.
    }
    \label{tab:against_others_lm}
  \end{minipage}%
  \hspace{.02\textwidth}%
  \begin{minipage}[c]{0.65\textwidth}
    \vspace{1mm}
\newcolumntype{R}{>{\raggedleft\arraybackslash}X}
\tablefontsize
\begin{tabularx}{1\linewidth}{Xrrrr}
    \toprule
    Algorithm 
    & \multicolumn{1}{l}{Test perplexity}
    & \multicolumn{1}{l}{Data/epoch} 
    & \multicolumn{2}{l}{Time per batch}\\
    \cmidrule(lr){1-1} \cmidrule(lr){2-2} \cmidrule(lr){3-3} \cmidrule(lr){4-5}SGD& 91 \tikz{
\draw[gray,line width=.3pt] (0,0) -- (1.2,0);
\draw[white, line width=0.01pt] (0,-2pt) -- (0,2pt);
\draw[black,line width=1pt] (0.060968834661668234,0) -- (0.06657085787865429,0);
\draw[black,line width=1pt] (0.060968834661668234,-2pt) -- (0.060968834661668234,2pt);
\draw[black,line width=1pt] (0.06657085787865429,-2pt) -- (0.06657085787865429,2pt);}
    & 7730 MB
    & 300 ms 
    & {\color{gray}$+0\%$}\\\signum& 142 \tikz{
\draw[gray,line width=.3pt] (0,0) -- (1.2,0);
\draw[white, line width=0.01pt] (0,-2pt) -- (0,2pt);
\draw[black,line width=1pt] (1.0256472679876512,0) -- (1.0421470789755545,0);
\draw[black,line width=1pt] (1.0256472679876512,-2pt) -- (1.0256472679876512,2pt);
\draw[black,line width=1pt] (1.0421470789755545,-2pt) -- (1.0421470789755545,2pt);}
    & 242 MB
    & 424 ms 
    & {\color{gray}$+41\%$}\\\textbf{\hspace{0pt}Rank 4\hspace{0pt}}& 91 \tikz{
\draw[gray,line width=.3pt] (0,0) -- (1.2,0);
\draw[white, line width=0.01pt] (0,-2pt) -- (0,2pt);
\draw[black,line width=1pt] (0.05309547670425913,0) -- (0.058987131426411286,0);
\draw[black,line width=1pt] (0.05309547670425913,-2pt) -- (0.05309547670425913,2pt);
\draw[black,line width=1pt] (0.058987131426411286,-2pt) -- (0.058987131426411286,2pt);}
    & 64 MB
    & 134 ms 
    & {\color{gray}$-55\%$}\\
    \bottomrule
\end{tabularx}
  \end{minipage}
\end{table}

\section{Conclusion}

Gradient compression is a promising approach to tackling the communication bottleneck in synchronous distributed optimization.
Thus far, however, it has not found widespread adoption because existing compression schemes either run slower than SGD with optimized all-reduce gradient aggregation, or more importantly do not reach the same test performance.
We see \powersgd\ as the first practical gradient compression method, and believe it is ready for adaptation in practice.

The key to the practicality of \powersgd\ is its linear compression scheme that is cheap to compute and allows for \allreduce\ gradient aggregation, while simultaneously matching the test performance of full-precision SGD.
This speedup gained over SGD actually \emph{increases} for larger models such as those commonly found in NLP.
Further, as a result of our modifications to the error feedback algorithm, \powersgd\ is a plug-in replacement for SGD with momentum, avoiding the need for additional hyper-parameter tuning. We expect that these properties of \powersgd\ will enable training of even larger models with even more workers than what is possible with full-precision SGD.

While \powersgd\ enables faster training with larger batch sizes, increasing batch sizes are known to eventually suffer from a `generalization gap'~\citep{shallue2018measuring}. This is an orthogonal issue that we see as the next step towards solving large-scale training.
In our experiments, we have observed that \powersgd\ can achieve higher test accuracy than SGD. Combined with the intriguing links between low-rankedness and generalization, this indicates that \powersgd\ may also be helpful for closing the generalization gap in large batch training.

\subsubsection*{Acknowledgements}

We thank Alp Yurtsever and Tao Lin for valuable discussions and the reviewers for their feedback.
This project was supported by SNSF grant $200021\_175796$, as well as a Google Focused Research Award.



\bibliography{papers}
\bibliographystyle{icml2019}
\onecolumn
\part*{Appendix}
\appendix

\section{Discussion of convergence} \label{sec:convergence_proof}
The proof of convergence of EF-SGD with momentum can be derived by incoporating a few key changes to the proof of \cite{karimireddy2019error}: i) we are in a multi-worker setting, and ii) we incorporate the techniques introduced by \cite{ghadimi2016accelerated} to handle the additional \emph{momentum}. Further, $\norm{\cdot}^2$ unless otherwise specified is always the standard euclidean norm for vectors, and is the \emph{Frobenius} norm for matrices.

Suppose that we want to minimize a continuous (possibly) non-convex function $f \colon \real^d \to \real$:
\[
	f^\star = \min_{\xx \in \real^d} f(\xx)\,.
\]
The classic stochastic gradient algorithm (SGD) \cite{robbins1951stochastic} when adapted to the distributed optimization setting performs iterations of the form
\begin{equation}\label{eqn:sgd}
  \xx_{t+1} := \xx_t - \gamma\, \gg_t\,, \text{ where}
\end{equation}
\[
	\gg_t = \frac{1}{W}\sum_{w=1}^W \gg_{t,w}
	\hspace{1cm}
	\text{and}
	\hspace{1cm}
	\expect\sbr{\gg_t} = \nabla f(\xx_t)\,.
\]
Here $\gamma \in \real$ is the step-size (or learning-rate) and $\gg_{t,w}$ is the stochastic gradient computed by the $w$th worker for $w \in \{1,\dots,W\}$ workers.

Now EF-SGD (Algorithm \ref{alg:ef_sgd}) when run on the $W$ workers with step-size $\gamma$ and momentum  parameter $\lambda$ can be rewritten making the dependence on iteration $t$ explicit as follows:

\begin{equation}\label{eqn:ef-sgd}
\begin{split}
	\Delta_t' &= \textsc{Decompress}(\textsc{compress}(\gg_t + \ee_t))\,,\\
	\mm_{t+1} &= \Delta_t' + \lambda \mm_t\,,\\
	\xx_{t+1} &= \xx_t - \gamma(\Delta_t' + \mm_{t+1}) \,, \text{ and}\\
	\ee_{t+1} &= (\gg_t + \ee_t) - \Delta_t' \,.
\end{split}
\end{equation}

\subsection{Eigen compression}\label{subsec:eigen}
\begin{assumption}[Eigen compression]\label{asm:eigen-compression}
Consider any matrix $M = g_t + e_t$ encountered during the run of Algorithm \ref{alg:ef_sgd} such that $M$ is of rank $R$. Further, suppose that $\cC_r(M)$ is the best rank-$r$ approximation of $M$ i.e.
\[
	\cC_r(M) = \argmin_{C}\norm{M - C}^2\,.
\]
Then we assume that there exists a $\delta_{\texttt{e}, r} > 0$ such that
\[
	\norm{M - \cC_r(M)}^2 \leq (1  - \delta_{\texttt{e}, r})\norm{M}^2\,\ a.s.
\]
\end{assumption}
We state the below standard fact from linear algebra.
\begin{remark}[Best rank-$r$ approximation]
	Suppose we are given a matrix $M$ of rank $n$ whose singular value decomposition is
	\[
		M = \sum_{i=1}^n \sigma_i \uu_i \vv_i^\top\,,
	\]
	where the singular-values $(\sigma_i)$ are sorted in descending order. Then the best rank-$r$ approximation of $M$ for $r \leq n$ is
	\[
	\cC_r(M) = (\sum_{i=1}^r \sigma_i \uu_i \vv_t^\top)Q\,,
	\]
	where $Q \in \real^{r\times r}$ is an orthogonal matrix, and further the quality of its approximation is bounded by
	\[
		\norm{M - \cC_r(M)}^2 = \rbr*{1 - \frac{\sum_{i=1}^r \sigma_i^2}{\sum_{i=1}^n \sigma_i^2}} \norm{M}^2\,.
	\]
\end{remark}
Thus if we used Algorithm \ref{alg:ef_sgd} with exact rank-$r$ approximation of the gradients, we would converge at rate dictated by the eigen-spectrum of the gradients. If the singular values are `top-heavy' \ie\ the largest $r$ values are significantly larger than the rest, then a rank-$r$ approximation is quite accurate. As demonstrated in \citep{wang2018atomo}, the eigen-spectrum of stochastic gradients in common deep learning tasks is indeed `top-heavy'. Thus we can expect $\delta_{\texttt{e}, r}$ to be bounded away from 0 even for very small $r$ (\eg\ 1 or 2). Of course computing the actual top eigenvectors of the stochastic gradients is very computationally expensive, and more-over is not linear (and hence does not support \emph{reduce}).

\subsection{Subspace iteration}\label{subsec:subspace}\mj{why call it subspace here and not power?}
The key innovation in \powersgd\ is to use only a \emph{single} step of subspace (or power) iteration to give a fast low rank approximation \citep{stewart1975methods} to the given matrix, which in our case is a stochastic gradient. However, a single step of subspace iteration in general does not result in an adequate low-rank approximation of the input matrix.
To combat this, and to at the same time reduce the variance of the stochastic gradient approximation compared to the full (deterministic) gradient, we propose the \emph{reuse} of the low-rank approximation from the previous iteration as the starting point for the current iteration.
This is in spite of the target matrices which are trying to approximate are \emph{changing}, as the parameters evolve.
Nevertheless, reuse here is justified because the full gradient does not change very fast (the gradient is Lipschitz by assumption) and we only perform a tiny update at each step, so can be assumed to be stationary within a small number of steps.
Intuitively, by linearity of the subspace operation, the sequence of subspace steps with the reuse then is converging to the eigenvector of the averaged stochastic gradients over these steps, thus having a lower variance than the analogue without re-use, which has no such averaging effect.
\mj{disregarding the effect of error-feedback for now for simplicity. can we explain this a bit in more detail and then have a short summary as a motivation for reuse in the main paper also?}

For simplicity, we assume all matrices to be square and symmetric in this sub-section. These insights can be generalized to arbitrary matrices but with a substantial increase in complexity of exposition. Here, we simply note that for any non-square matrix $A$, we can instead consider
\[
\tilde A = \begin{bmatrix}
    0       & A \\
    A^\top   & 0
\end{bmatrix}
\]
which is symmetric and has the same eigenvectors and eigenvalues as the original matrix $A$---see \cite{stewart1976simultaneous} for more details on handling such cases.

We can now state an informal theorem about the convergence of subspace iteration.
\begin{theorem}\label{thm:subspace-converge}
Suppose that we run subspace iteration as in \eqref{eqn:subspace-iter} on a fixed matrix $A_t = M$. Also let $M = \sum_{i=1}^n \sigma_i \uu_i\uu_i^\top$ be the eigen decomposition of $M$ with $\sigma_1 \geq \dots \sigma_r > \sigma_{r+1} \geq \dots \geq \sigma_n$. Then there exists an orthogonal matrix $Q \in \real^{r \times r}$ such that
\[
 \lim_{t = \infty} X_t = [\uu_1,\dots,\uu_r]Q\,.
 \]
 In other words, \eqref{eqn:subspace-iter} recovers the best rank-$r$ approximation of $M$ as long as there is a gap between the $\sigma_r$ and $\sigma_{r+1}$ eigenvalues.
\end{theorem}
Suppose that at each iteration we receive a matrix $A_t \in \real^{n \times n}$ whose expectation is the same fixed matrix $M \in \real^{n \times n}$.
Starting from an orthonormalized $X_0 \in \real^{n \times r}$ (\ie\ $X_0^\top X_0 = I_r$), the rank-$r$ subspace iteration algorithm performs the following update:
\begin{equation}\label{eqn:subspace-iter}
	X_{t+1} = \textsc{orthogonalize}(A_t X_t)\,.
\end{equation}
The final output of the algorithm (i.e.) the matrix approximation is $(A_{T+1}X_T)  X_T^\top$. This closely resembles the method of \powersgd\ as outlines in Algorithm \ref{alg:powersgd}. We recommend \citep{arbenz2012lecture} for an in-depth analysis of the (non-varying) subspace iteration algorithm.

\begin{remark}[Orthogonalization is a linear operation]\label{rem:qr}
We recall some more facts from linear algebra. For any square matrix $B$, there exists an orthogonal matrix $Q$ and a triangular matrix $R$ such that $QQ^\top = I$ and $B = QR$. This is true \eg\ if we use Gram–Schmidt procedure to ortho-normalize $B$: Suppose $\textsc{orthogonalize}(B)$ uses the Gram–Schmidt procedure to orthogonalize $B$. Then there exists a triangular matrix $R$ such that
\[
	\textsc{orthogonalize}(B) = B R^{-1}\,.
\]
\end{remark}

\begin{proof}
It is easy to see that for any orthogonal matrix $Q$, the matrix $[\uu_1,\dots,\uu_r]Q$ is also orthogonal, and further is the fixed point of \eqref{eqn:subspace-iter}. In fact all rank-$r$ matrices which are fixed points of \eqref{eqn:subspace-iter} are of this form.

We will use the observation in Remark \ref{rem:qr} to rewrite the update \eqref{eqn:subspace-iter} in a more convient fashion. There exist tringular matrices $R_0, \dots, R_t$ such that
\[
	X_{t+1} = \textsc{orthogonalize}(A_t X_t) = A_t X_t R_t^{-1} = (A_t A_{t-1} \cdots A_0)X_0 (R_0^{-1} R_1^{-1} \cdots R_t^{-1})\,.
\]
Thus $X_{t+1}$ can alternatively be written as
\[
	X_{t+1} = \textsc{orthogonalize}((A_t A_{t-1} \cdots A_0)X_0) =\textsc{orthogonalize}(M^{t+1} X_0) \,.
\]
Here we assumed that the matrix was fixed i.e. $A_t = M$. Let us further assume that $X_0$ has a non-zero support on the first $r$ eigenvectors of $M$. Then, a gap in the eigenvalues $\sigma_r > \sigma_{r+1}$ implies that $\textsc{orthogonalize}(M^{t+1} X_0)$ converges to $[\uu_1,\dots,\uu_r]Q$. We refer to Chapter 7.2 of \cite{arbenz2012lecture} for the actual proof of this fact.
\end{proof}

\subsection{Single/multi worker equivalence}\label{subsec:single-multi}
The difference between the update as written in \eqref{eqn:ef-sgd} and Algorithm \ref{alg:ef_sgd} is that the error computation and compression is performed on the \emph{aggregated} gradient $\gg_t$ instead of on the individual workers' gradients $\gg_{t,w}$. While in general these are not equivalent, the linearity of \powersgd\ ensures that these are indeed equivalent. This implies that \powersgd\ has the neat property that the algorithm is equivalent if run on $W$ workers or a single worker with a larger batch-size. This does not hold for most other schemes (e.g. sign based compression schemes, QSGD, etc.).
\begin{lemma}[Equivalence of single worker and multi worker updates] The updates in \powersgd\ (\ie\ Algorithm \ref{alg:ef_sgd} using Compressor \ref{alg:powersgd}) are equivalent to the updates \eqref{eqn:ef-sgd}.
\end{lemma} \label{lem:single-multi-equal}
\begin{proof}
	Consider the update performed by \powersgd\ for abrtiary vectors $\{\vv_w\}$. Let $\cC(\vv_w)$ be the compressed version of $\vv_w$ for $w \in \{1,\dots,W\}$. Then by design of \powersgd\,, the following holds:
	\[
		\textsc{Decompress}(\textsc{aggregate}(\cC(\vv_1), \dots, \cC(\vv_W))) = \textsc{Decompress}(\cC(\frac{1}{W}\sum_{w} \vv_w))\,.
	\]
	This implies that running the algorithm on multiple workers, or running it on a single worker with a larger batch-size is identical. In particular,
	\begin{align*}
		\textsc{Decompress}(\textsc{aggregate}(\cC(\gg_{t,1} + \ee_{t,1}), \dots, \cC(\gg_{t,W} + \ee_{t,W}))) &\\
		&\hspace*{-5cm}= \textsc{Decompress}(\cC(\frac{1}{W}\sum_{w} \gg_{t,w} + \ee_{t,w}))\\
		&\hspace*{-5cm}= \textsc{Decompress}(\frac{1}{W}\cC(\gg_t + \ee_t))\,.
	\end{align*}
\end{proof}

\raggedbottom
\pagebreak

\section{Cluster specifications} \label{sec:cluster_specifications}

\begin{itemize}[leftmargin=*]
    \item 8 nodes
    \item GPUs: $2 \times$ Nvidia GeForce GTX Titan X with 12 GB memory per node
    \item GPU connection: traversing PCIe and the SMP interconnect between NUMA nodes
    \item CPU: Intel Xeon E5-2680 v3 @ 2.50Ghz, 48 cores
    \item System memory: 251GiB
    \item Ethernet: 10Gbit/s SFI/SFP+
    \item \emph{Fat tree} network topology
    \item Runing \pytorch\ 1.1 on Anaconda Python 3.7
\end{itemize}

\textbf{Timings of collective communication operations}

The figure below shows timings for the \nccl\ backend, which is the default in our experiments, and the \gloo\ backend. Note that \nccl\ does not support the `gather' operation in \pytorch at the time of writing.

\includegraphics[width=0.5\textwidth]{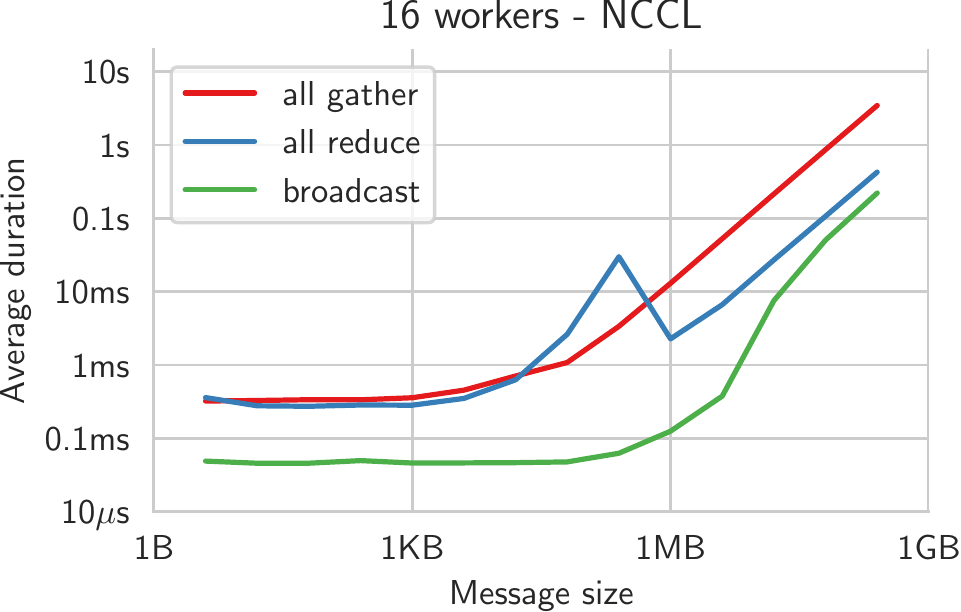}
\includegraphics[width=0.5\textwidth]{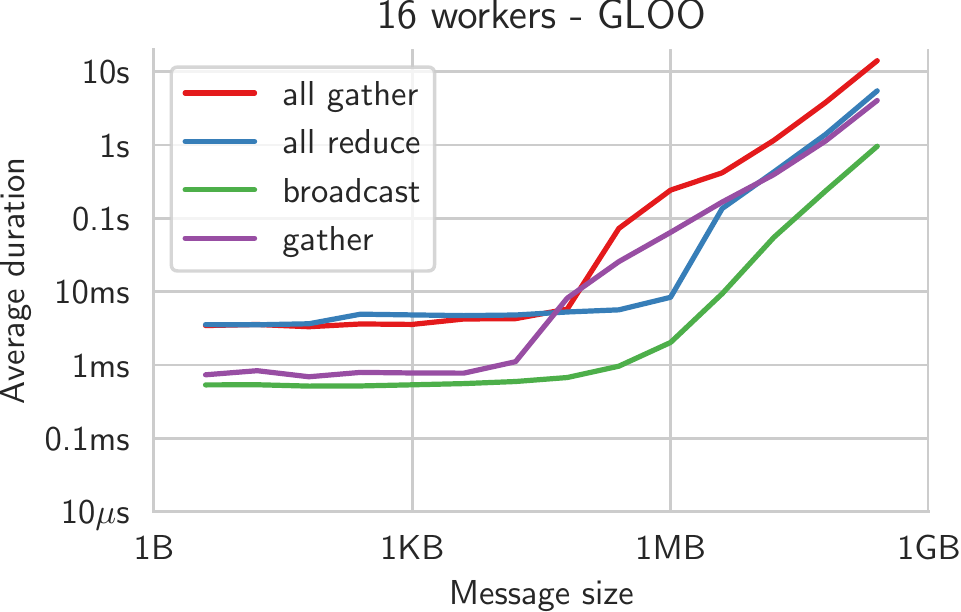}

\raggedbottom
\pagebreak

\section{Convergence curves}\label{app:convergence_curves}
\begin{figure}[H]
  \begin{subfigure}[t]{0.5\textwidth}
  \begin{minipage}{\textwidth}\hspace{.8cm}\centering Image classification on \cifar\end{minipage}
    \includegraphics[width=\textwidth]{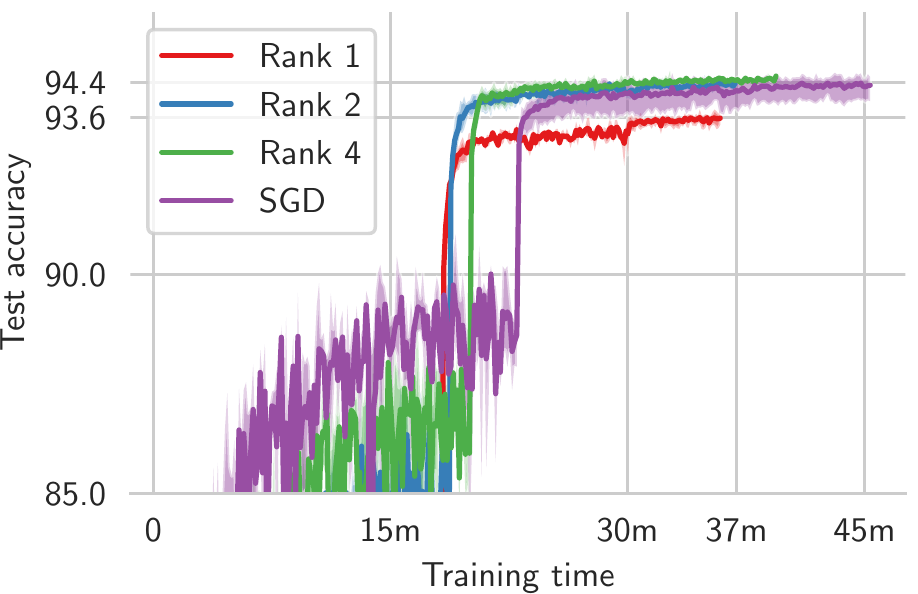}
  \end{subfigure}
  \begin{subfigure}[t]{0.5\textwidth}
  \begin{minipage}{\textwidth}\hspace{.8cm}\centering Language modeling with \wikitext\end{minipage}
    \includegraphics[width=\textwidth]{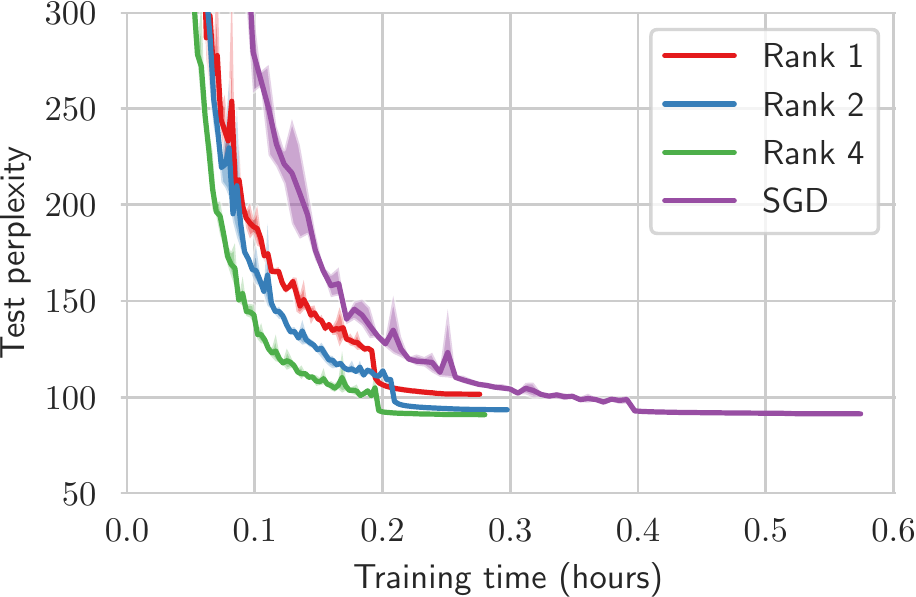}
  \end{subfigure}\\[8pt]
  \begin{subfigure}[t]{0.5\textwidth}
  \begin{minipage}{\textwidth}\hspace{.8cm}\centering Image classification on \cifar\end{minipage}
    \includegraphics[width=\textwidth]{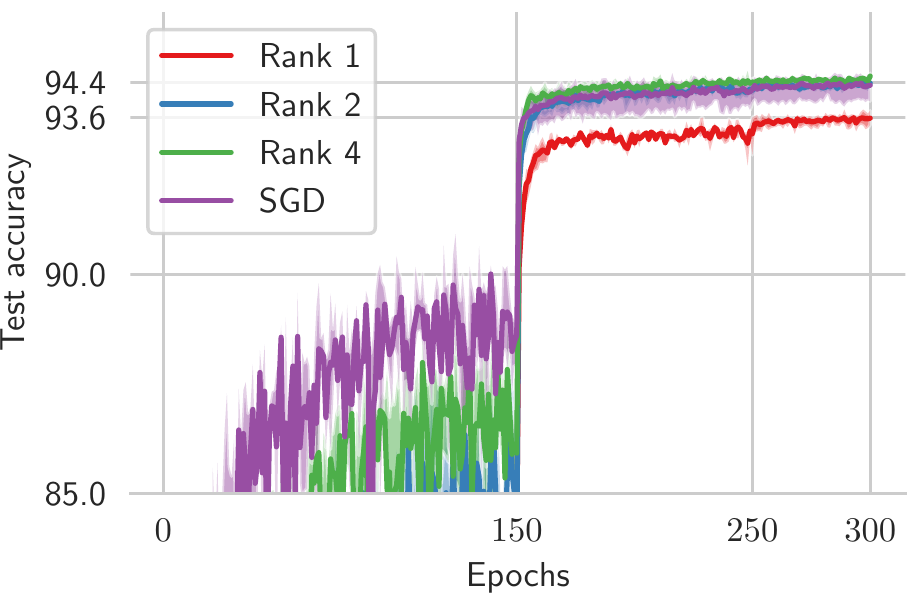}
  \end{subfigure}
  \begin{subfigure}[t]{0.5\textwidth}
  \begin{minipage}{\textwidth}\hspace{.8cm}\centering Language modeling with \wikitext\end{minipage}
    \includegraphics[width=\textwidth]{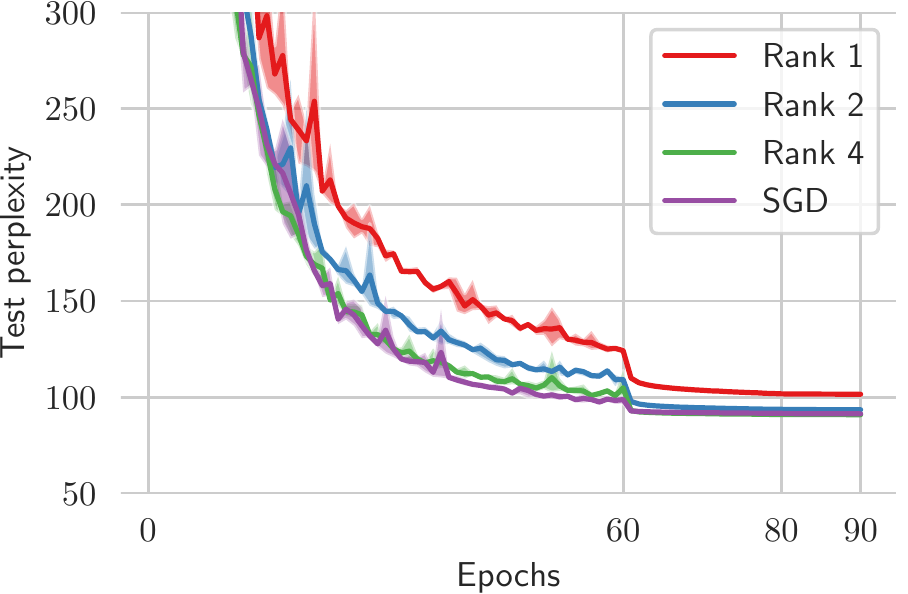}
  \end{subfigure}
  \caption{Convergence curves of \powersgd\ with varying rank. This figure is meant to give context to the final results and timings presented in Table~\ref{tab:which_rank}. In two different tasks, \powersgd\ with high enough rank can achieve the test quality of full-precision SGD with lower wall-clock duration. Contrary to Table~\ref{tab:which_rank}, these timings include testing overhead at the end of each epoch, checkpointing, and other bookkeeping. Shaded areas show the min---max values over 3 replications of the experiments.}
\end{figure}

\begin{figure}[H]
  \begin{subfigure}[t]{0.5\textwidth}
  \begin{minipage}{\textwidth}\hspace{.8cm}\centering Image classification on \cifar\end{minipage}
    \includegraphics[width=\textwidth]{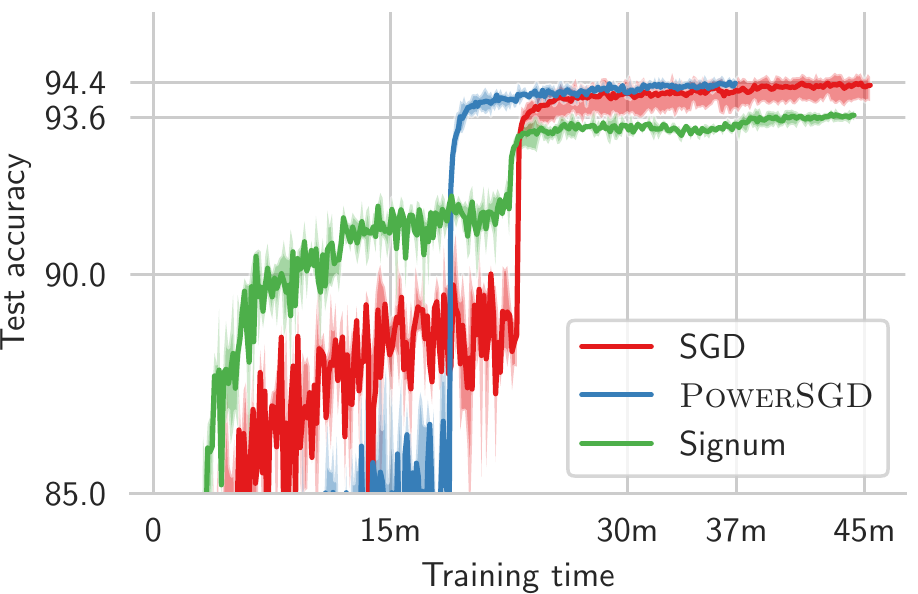}
  \end{subfigure}
  \begin{subfigure}[t]{0.5\textwidth}
  \begin{minipage}{\textwidth}\hspace{.8cm}\centering Language modeling with \wikitext\end{minipage}
    \includegraphics[width=\textwidth]{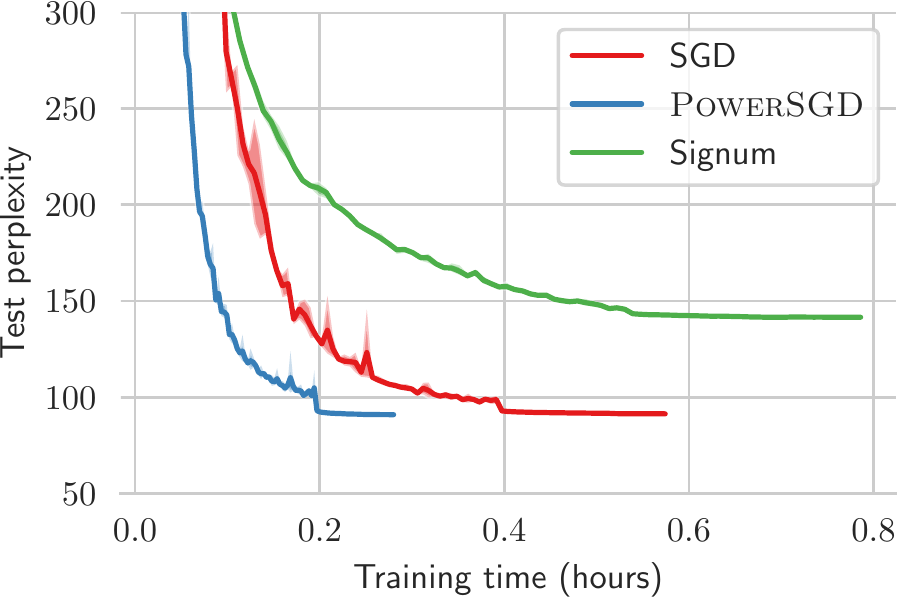}
  \end{subfigure}\\[8pt]
  \begin{subfigure}[t]{0.5\textwidth}
  \begin{minipage}{\textwidth}\hspace{.8cm}\centering Image classification on \cifar\end{minipage}
    \includegraphics[width=\textwidth]{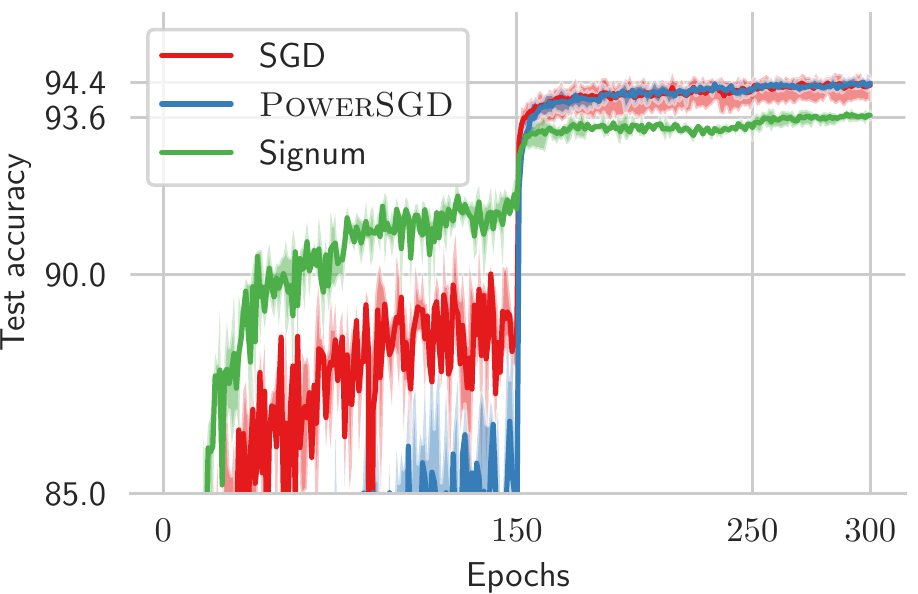}
  \end{subfigure}
  \begin{subfigure}[t]{0.5\textwidth}
  \begin{minipage}{\textwidth}\hspace{.8cm}\centering Language modeling with \wikitext\end{minipage}
    \includegraphics[width=\textwidth]{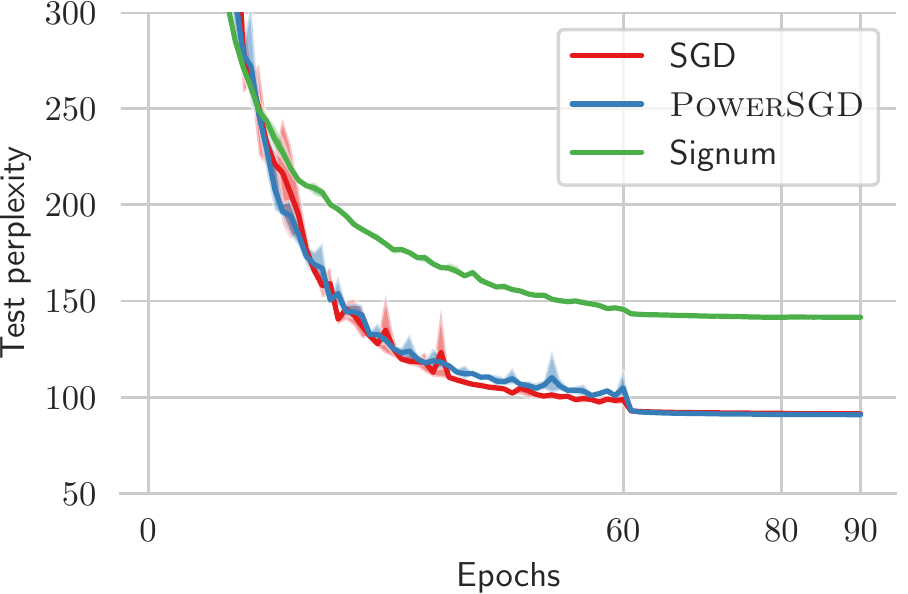}
  \end{subfigure}
  \caption{Convergence curves comparing \powersgd\ to the \signum\ optimizer~\cite{bernstein2019iclr} (with tuned learning rate).
  Out of the compared methods, \signum\ came out as the most competitive.
  This figure is meant to give context to the final results and timings presented in Table~\ref{tab:against_others}. Contrary to Table~\ref{tab:which_rank}, these timings include testing overhead at the end of each epoch, checkpointing, and other bookkeeping. Shaded areas show the min---max values over 3 replications of the experiments.}
\end{figure}

\raggedbottom
\pagebreak

\section{Language Modeling with Transformers}\label{app:fairseq}

In this case study, we assess PowerSGD's universality and ease of tuning.
We implemented PowerSGD communication in Facebook AI Research's \texttt{fairseq} library~\citep{ott2019fairseq}.
We trained fairseq's language modeling example\footnote{\url{https://github.com/pytorch/fairseq/tree/920b85d4bd39e181229db5639c701c854c83ec5c/examples/language_model}} with transformers~\citep{baevski2019adaptive} on Google's public cloud. The communication infrastructure, hardware, number of workers (32), and model architecture are all different from any experiments we have conducted before. See Table~\ref{tab:fairseq-defaults} for details.

\begin{figure}[H]
  \begin{subfigure}[t]{0.5\textwidth}
  \begin{minipage}{\textwidth}\hspace{.8cm}\centering Wall-clock time\end{minipage}
    \includegraphics[width=\textwidth]{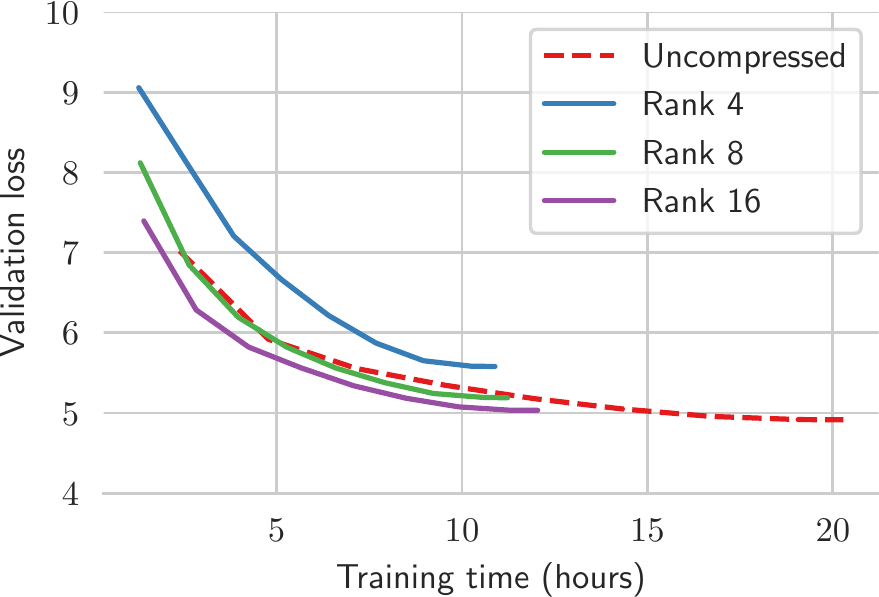}
  \end{subfigure}
  \begin{subfigure}[t]{0.5\textwidth}
  \begin{minipage}{\textwidth}\hspace{.8cm}\centering Update steps\end{minipage}
    \includegraphics[width=\textwidth]{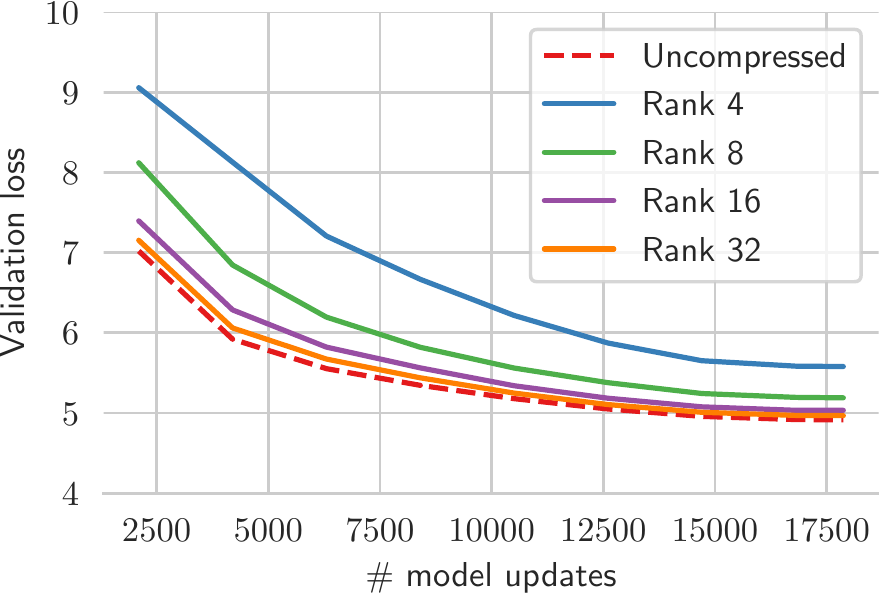}
  \end{subfigure}
  \caption{
    Language Modeling on \wikitext\ with Transformers.
    With a large enough rank, \powersgd\ can roughly match the validation loss of full-precision SGD in the same number of iterations.
    A speedup of $1.5\times$ in time-to-accuracy (loss=5) is achieved with a rank of 16.
  }
  \label{fig:fairseq_curves}
\end{figure}

The results of our experiments for various ranks are shown in Figure~\ref{fig:fairseq_curves} and Table~\ref{tab:fairseq_results}.
For this task, we need a higher rank than previously (32 vs 4) to achieve a validation loss comptetitive to uncompressed SGD.
We hypothesize this may be due differences in architecture to the cosine learning rate schedule.
Nevertheless, even at this higher rank, we achieve a time-to-accuracy (to $\text{loss}=5$) of around $1.5\times$ and a compression ratio of $14\times$.
These numbers could probably be further improved by re-tuning learning-rate-related hyperparameters.

\begin{table}[h]%
  \scriptsize%
  \caption{Experimental setting for the experiments in Appendix~\ref{app:fairseq}}
  \vspace{1mm}
  \label{tab:fairseq-defaults}%
  \begin{tabularx}{\linewidth}{lX}
      \toprule
      Dataset & WikiText-103 \\
      Architecture & Transformer-based~\citep{baevski2019adaptive} \\
      Framework \& defaults & \url{https://github.com/pytorch/fairseq/tree/920b85d4bd39e181229db5639c701c854c83ec5c/examples/language_model} \\
      \midrule
      Number of workers & 32 \\
      Backend & \nccl\ (fastest in \pytorch) \\
      Hardware & n1-standard-8 nodes on Google Cloud with 1 Nvidia Tesla K80 GPU \\
      \midrule
      Hyperparameters & Taken from the example, not re-tuned, \\
      & with minor changes for the higher number of workers and different GPU memory: \\
      \texttt{lr period updates} & 16875 \\
      \texttt{max update} & 17875 \\
      \texttt{max tokens (valid)} & 1536 (to fit on a K80 gpu) \\
      \texttt{tokens per sample} & 1536 (to fit on a K80 gpu) \\
      \texttt{warmup updates} & 1000 \\
      \texttt{update freq} & $\left[1\right]$ --- don't aggregate multiple mini-batches locally \\
      \midrule
      Optimizer & original: Nesterov accelerated gradient, we just added PowerSGD for communication \\
      Learning rate & original cosine schedule from the example \\
      \midrule
      Float precision & 32-bit (16-bit is unavailable on the K80) \\
      \midrule
      Repetitions & 1 \\
      \bottomrule
  \end{tabularx}
\end{table}

\begin{table}[H]
  \centering
  \begin{minipage}{\textwidth}
    \DeclareRobustCommand\dotOne{\tikz{\fill[fill=color1] (0,0) rectangle (5pt,5pt);}~}
    \DeclareRobustCommand\dotTwo{\tikz{\fill[fill=color2] (0,0) rectangle (5pt,5pt);}~}
    \DeclareRobustCommand\dotThree{\tikz{\fill[fill=color3] (0,0) rectangle (5pt,5pt);}~}
    \DeclareRobustCommand\dotThreeStriped{\tikz{\fill[fill=color3] (0,0) rectangle (5pt,5pt);\fill[pattern=north east lines, pattern color=black!70!color3] (0,0) rectangle (5pt,5pt);}~}
    \caption{
      \powersgd\ for Language Modeling with Transformers. With rank 32, \powersgd\ achieves similar validation loss to uncompressed SGD in the same number of update steps.
      At this rank, the compression ratio is $14\times$ and we can train the model in 12h compared to 20h for the baseline.
      }
      \label{tab:fairseq_results}
      \vspace{1mm}
\newcolumntype{R}{>{\raggedleft\arraybackslash}X}
\tablefontsize
\begin{tabularx}{1\linewidth}{XllXX}
    \toprule 
    Compression & Total training time & & Compression ratio & Validation loss \\
    & \tiny for 17875 updates &&& \tiny at 17875 updates \\
    \midrule
        Uncompressed & 
        \tikz{
        \fill[fill=color1] (0.0,0) rectangle (0.3306213454312762,0.2);
        \fill[fill=color2] (0.3306213454312762,0) rectangle (0.8769635120239601,0.2);
        \fill[fill=color3] (0.8769635120239601,0) rectangle (2.135864031798971,0.2);        
    } &
        20h &
        $1\times$ &
        4.92 \\
        Rank 4 & 
        \tikz{
        \fill[fill=color1] (0.0,0) rectangle (0.3459546783619434,0.2);
        \fill[fill=color2] (0.3459546783619434,0) rectangle (0.9151796831283344,0.2);
        \fill[fill=color3] (0.9151796831283344,0) rectangle (1.0997569381270182,0.2);        
    } &
        11h &
        $105\times$ &
        5.58 \\
        Rank 8 & 
        \tikz{
        \fill[fill=color1] (0.0,0) rectangle (0.35053165209823267,0.2);
        \fill[fill=color2] (0.35053165209823267,0) rectangle (0.8884816923075287,0.2);
        \fill[fill=color3] (0.8884816923075287,0) rectangle (1.137265134901474,0.2);        
    } &
        11h &
        $55\times$ &
        5.19 \\
        Rank 16 & 
        \tikz{
        \fill[fill=color1] (0.0,0) rectangle (0.3545320610741785,0.2);
        \fill[fill=color2] (0.3545320610741785,0) rectangle (0.9382863735232084,0.2);
        \fill[fill=color3] (0.9382863735232084,0) rectangle (1.2282298905319653,0.2);        
    } &
        12h &
        $28\times$ &
        5.03 \\
        Rank 32 & 
        \tikz{
        \fill[fill=color1] (0.0,0) rectangle (0.32692591732541687,0.2);
        \fill[fill=color2] (0.32692591732541687,0) rectangle (0.8669783226801236,0.2);
        \fill[fill=color3] (0.8669783226801236,0) rectangle (1.323730565489659,0.2);        
    } &
        13h &
        $14\times$ &
        4.97 \\
    & \tikz{
        \draw[black] (0,0) -- (2.2,0);
        \draw[black] (0.0,-2pt) -- (0.0,2pt);
        \draw[black] (0.44000000000000006,-2pt) -- (0.44000000000000006,2pt) node[anchor=north,yshift=-0.75mm] {\tiny$4$h};
        \draw[black] (0.8800000000000001,-2pt) -- (0.8800000000000001,2pt) node[anchor=north,yshift=-0.75mm] {\tiny$8$h};
        \draw[black] (1.3200000000000003,-2pt) -- (1.3200000000000003,2pt) node[anchor=north,yshift=-0.75mm] {\tiny$12$h};
        \draw[black] (1.7600000000000002,-2pt) -- (1.7600000000000002,2pt) node[anchor=north,yshift=-0.75mm] {\tiny$16$h};
        \draw[black] (2.2,-2pt) -- (2.2,2pt) node[anchor=north,yshift=-0.75mm] {\tiny$20$h};
    }\vspace{-1mm} \\
    \bottomrule
\end{tabularx} \\
      \dotOne Forward pass \hspace{1mm}
      \dotTwo Backward pass \hspace{1mm}
      \dotThree Gradient exchange including computation
  \end{minipage}
\end{table}

\vfill
\section{The need for error feedback}\label{sec:need-error-feedback}

\begin{figure}[H]
  \centering
  \begin{subfigure}[t]{0.5\textwidth}
  \begin{minipage}{\textwidth}\hspace{.8cm}\centering \resnet\ on \cifar\end{minipage}
    \includegraphics[width=\textwidth]{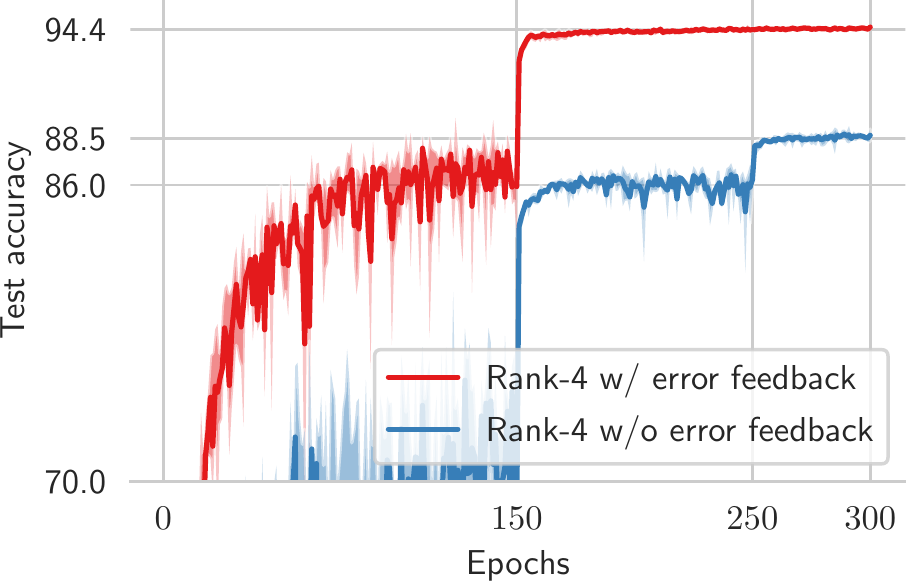}
  \end{subfigure}
  \caption{
    PowerSGD with and without error feedback compared.
    While rank-4 \powersgd\ achieves the same test accuracy as full-precision SGD,
    the same method without error feedback does not converge to a good accuracy at all.
    Both experiments use the same learning rate that was tuned for full-precision SGD.
  }
  \label{fig:need-error-feedback}
\end{figure}

\raggedbottom
\pagebreak
\section{Network parameters}\label{sec:network-parameters}

See Table~\ref{tab:ResNet18_parameters} and Table~\ref{tab:LSTM_parameters} for an overview of parameters in the models used.

\begin{table}[h]
\caption{Parameters in the ResNet18 architecture and their shapes. The table shows the per-tensor compression ratio achieved by rank-$r$ \powersgd.}
\label{tab:ResNet18_parameters}
\vspace{1mm}
\small%
\newcolumntype{R}{>{\raggedleft\arraybackslash}X}
\begin{tabularx}{\linewidth}{Xllrr}
\toprule
Parameter
& Gradient tensor shape
& Matrix shape
& Uncompressed
& Compression \\
\cmidrule(lr){1-5}
layer4.1.conv2 
& $512 \times 512 \times 3 \times 3$
& $512 \times 4608$
& 9216 KB
& $461/r \; \times$ \\
layer4.0.conv2 
& $512 \times 512 \times 3 \times 3$
& $512 \times 4608$
& 9216 KB
& $461/r \; \times$ \\
layer4.1.conv1 
& $512 \times 512 \times 3 \times 3$
& $512 \times 4608$
& 9216 KB
& $461/r \; \times$ \\
layer4.0.conv1 
& $512 \times 256 \times 3 \times 3$
& $512 \times 2304$
& 4608 KB
& $419/r \; \times$ \\
layer3.1.conv2 
& $256 \times 256 \times 3 \times 3$
& $256 \times 2304$
& 2304 KB
& $230/r \; \times$ \\
layer3.1.conv1 
& $256 \times 256 \times 3 \times 3$
& $256 \times 2304$
& 2304 KB
& $230/r \; \times$ \\
layer3.0.conv2 
& $256 \times 256 \times 3 \times 3$
& $256 \times 2304$
& 2304 KB
& $230/r \; \times$ \\
layer3.0.conv1 
& $256 \times 128 \times 3 \times 3$
& $256 \times 1152$
& 1152 KB
& $209/r \; \times$ \\
layer2.1.conv2 
& $128 \times 128 \times 3 \times 3$
& $128 \times 1152$
& 576 KB
& $115/r \; \times$ \\
layer2.1.conv1 
& $128 \times 128 \times 3 \times 3$
& $128 \times 1152$
& 576 KB
& $115/r \; \times$ \\
layer2.0.conv2 
& $128 \times 128 \times 3 \times 3$
& $128 \times 1152$
& 576 KB
& $115/r \; \times$ \\
layer4.0.shortcut.0 
& $512 \times 256 \times 1 \times 1$
& $512 \times 256$
& 512 KB
& $171/r \; \times$ \\
layer2.0.conv1 
& $128 \times 64 \times 3 \times 3$
& $128 \times 576$
& 288 KB
& $105/r \; \times$ \\
layer1.1.conv1 
& $64 \times 64 \times 3 \times 3$
& $64 \times 576$
& 144 KB
& $58/r \; \times$ \\
layer1.1.conv2 
& $64 \times 64 \times 3 \times 3$
& $64 \times 576$
& 144 KB
& $58/r \; \times$ \\
layer1.0.conv2 
& $64 \times 64 \times 3 \times 3$
& $64 \times 576$
& 144 KB
& $58/r \; \times$ \\
layer1.0.conv1 
& $64 \times 64 \times 3 \times 3$
& $64 \times 576$
& 144 KB
& $58/r \; \times$ \\
layer3.0.shortcut.0 
& $256 \times 128 \times 1 \times 1$
& $256 \times 128$
& 128 KB
& $85/r \; \times$ \\
layer2.0.shortcut.0 
& $128 \times 64 \times 1 \times 1$
& $128 \times 64$
& 32 KB
& $43/r \; \times$ \\
linear 
& $10 \times 512$
& $10 \times 512$
& 20 KB
& $10/r \; \times$ \\
conv1 
& $64 \times 3 \times 3 \times 3$
& $64 \times 27$
& 7 KB
& $19/r \; \times$ \\
Bias vectors (total)
&
&
& 38 KB
& None \\
\cmidrule(lr){1-5}
\textbf{Total}
&
&
& 43 MB
& $243/r \; \times$ \\
\bottomrule
\end{tabularx}
\end{table}

\begin{table}[h]
\caption{Parameters in the LSTM architecture and their shapes. The table shows the per-tensor compression ratio achieved by rank-$r$ \powersgd.}
\label{tab:LSTM_parameters}
\vspace{1mm}
\small%
\newcolumntype{R}{>{\raggedleft\arraybackslash}X}
\begin{tabularx}{\linewidth}{Xllrr}
\toprule
Parameter
& Gradient tensor shape
& Matrix shape
& Uncompressed
& Compression \\
\cmidrule(lr){1-5}
encoder 
& $28869 \times 650$
& $28869 \times 650$
& 73300 KB
& $636/r \; \times$ \\
rnn-ih-l0 
& $2600 \times 650$
& $2600 \times 650$
& 6602 KB
& $520/r \; \times$ \\
rnn-hh-l0 
& $2600 \times 650$
& $2600 \times 650$
& 6602 KB
& $520/r \; \times$ \\
rnn-ih-l1 
& $2600 \times 650$
& $2600 \times 650$
& 6602 KB
& $520/r \; \times$ \\
rnn-hh-l1 
& $2600 \times 650$
& $2600 \times 650$
& 6602 KB
& $520/r \; \times$ \\
rnn-ih-l2 
& $2600 \times 650$
& $2600 \times 650$
& 6602 KB
& $520/r \; \times$ \\
rnn-hh-l2 
& $2600 \times 650$
& $2600 \times 650$
& 6602 KB
& $520/r \; \times$ \\
Bias vectors (total)
&
&
& 174 KB
& None \\
\cmidrule(lr){1-5}
\textbf{Total}
&
&
& 110 MB
& $310/r \; \times$ \\
\bottomrule
\end{tabularx}
\end{table}

\raggedbottom
\pagebreak
\section{Compressor implementation details}
\label{sec:implementations}

\subsection{Random Block}
This implements compression for error-feedback with momentum (Algorithm~\ref{alg:ef_sgd}).
\begin{algorithm}[H]
  \caption{Random Block compression}\label{alg:random_block}
  \begin{algorithmic}[1]
  \Function{compress}{update matrix $M \in \R^{n \times m}$}
    \State Treat $M$ as a vector of length $nm$.
    \State Sample an index $s$ uniformly between 0 and $nm-1$, using the same seed on all workers.
    \State The block length $b$ is set to $(m + n) r$ to match rank-$r$ \powersgd.
    \State \Return A consequtive memory slice $S=M(s:s+b)$.
  \EndFunction
  \Function{aggregate+decompress}{worker's slices $S_1\ldots S_W$}
  \State $\hat M \leftarrow \0 \in \R^{n \times m}$
  \State $\hat M(s:s+b) \leftarrow \frac{1}{W} \sum_{i=1}^W S_i$
    \Comment using \allreduce
  \State \Return $\hat M$
\EndFunction
  \end{algorithmic}
\end{algorithm}

\subsection{Random K}
This implements compression for error-feedback with momentum (Algorithm~\ref{alg:ef_sgd}).
\begin{algorithm}[H]
  \caption{Random $K$ compression}\label{alg:random_k}
  \begin{algorithmic}[1]
  \Function{compress}{update matrix $M \in \R^{n \times m}$}
    \State Treat $M$ as a vector of length $nm$.
    \State The number of samples $b$ is set to $(m + n) r$ to match rank-$r$ \powersgd.
    \State Sample a set of $b$ indices $I$ without replacement, using the same seed on all workers.
    \State \Return Looked up values $S=M(I)$.
  \EndFunction
  \Function{aggregate+decompress}{worker's values $S_1\ldots S_W$}
    \State $\hat M \leftarrow \0 \in \R^{n \times m}$
    \State $\hat M(I) \leftarrow \frac{1}{W} \sum_{i=1}^W S_i$
      \Comment using \allreduce
    \State \Return $\hat M$
  \EndFunction
  \end{algorithmic}
\end{algorithm}

\paragraph{Sampling of indices} We sample random indices on the CPU using Numpy. This operation is relatively expensive. Together with the many random lookups, this explains why Random $K$ compression is significantly slower than Random Block compression.

\subsection{Sign+Norm}
This implements compression for error-feedback with momentum (Algorithm~\ref{alg:ef_sgd}).
\begin{algorithm}[H]
  \caption{Sign$+$Norm compression}\label{alg:sign_and_norm}
  \begin{algorithmic}[1]
  \Function{compress}{update matrix $M \in \R^{n \times m}$}
    \State Compute the signs $S \in \left\{-1,1\right\}^{n \times m}$ of $M$
    \State Compute the $L_1$ norm $\ell$ of $M$.
    \State \Return ($\ell$, $S$)
  \EndFunction
  \Function{aggregate+decompress}{worker's norms $\ell_1\ldots \ell_W$ and signs $S_1 \ldots S_W$}
    \State \Return $\frac{1}{W} \sum_{i=1}^W \frac{\ell_i}{nm} S_i$
      \Comment Executed on all workers using \nccl's \allgather
  \EndFunction
  \end{algorithmic}
\end{algorithm}

Because \pytorch\ does not natively support data types smaller than 8 bits per scalar, we use a C++ extension~\citep{bernstein2019iclr} to actually send single bits to other workers. The employed \allgather\ operation from \nccl\ is faster than aggregation using a parameter server using \gloo. We cannot implement a parameter server in \nccl\ due to lack of a `gather' operation.

\subsection{Top K}
This implements compression for error-feedback with momentum (Algorithm~\ref{alg:ef_sgd}).
\begin{algorithm}[H]
  \caption{Top $K$ compression}\label{alg:top_k}
  \begin{algorithmic}[1]
    \Function{compress}{update matrix $M \in \R^{n \times m}$}
    \State Treat $M$ as a vector of length $nm$.
    \State The number of samples $b$ is set to $(m + n) r$ to match rank-$r$ \powersgd.
    \State Construct a list of $b$ indices $I$ corresponding to the top absolute values in $M$.
    \State \Return Looked up values $S=M(I)$ and indices $I$.
  \EndFunction
  \Function{aggregate+decompress}{worker's values $S_1\ldots S_W$ and indices $I_1 \ldots I_W$}
    \State $\hat M \leftarrow \0 \in \R^{n \times m}$
    \For{worker index $i$ in $1,\ldots,W$}
      \State $\hat M(I_i) \leftarrow \frac{1}{W} S_i$
        \Comment using \allgather\ in \nccl
    \EndFor
    \State \Return $\hat M$
  \EndFunction
  \end{algorithmic}
\end{algorithm}

The employed \allgather\ operation from \nccl\ is faster than aggregation using a parameter server using \gloo. We cannot implement a parameter server in \nccl\ due to lack of a `gather' operation.

\subsection{Signum}
This is our implementation of the \signum\ compression algorithm by \cite{bernstein2019iclr}. We run it in its original form, without error feedback, with momentum of 0.9, and a learning rate tuned based on 5 experiments in the 16-worker setting.
\begin{algorithm}[H]
  \caption{\signum\ compression}\label{alg:signum}
  \begin{algorithmic}[1]
  \Function{compress}{update matrix $M \in \R^{n \times m}$}
    \State Compute the signs $S \in \left\{-1,1\right\}^{n \times m}$ of $M$
    \State \Return $S$
  \EndFunction
  \Function{aggregate+decompress}{worker's signs $S_1 \ldots S_W$}
    \State \Return $\textsc{sign}(\sum_{i=1}^W S_i)$
      \Comment Majority vote, on all workers using \nccl's \allgather
  \EndFunction
  \end{algorithmic}
\end{algorithm}

Because \pytorch\ does not natively support data types smaller than 8 bits per number, we use a C++ extension~\cite{github2019signsgd} to actually send single bits to other workers. The employed \allgather\ operation from \nccl\ is faster than aggregation using a parameter server using \gloo. We cannot implement a parameter server in \nccl\ due to lack of a `gather' operation.

\subsection{Atomo}
This is our implementation of the Spectral Atomo algorithm presented by \cite{wang2018atomo}. We run it in its original form, without error feedback, with momentum of 0.9, and a learning rate tuned based on 4 experiments in the 16-worker setting.

\paragraph{Matix shape} Atomo differs from \powersgd\ in how it treats tensors as matrices. This results in lower compression at the same rank.

\paragraph{Number of sampled components} Atomo decomposes gradient matrices $M$ using a Singular Value Decomposition into $M \sim \sum_i U_{i:} S_{ii} V_{i:}^\top$ and importance-samples components from this summation based on probabilities derived from the absolute singular values $S_{ii}$. The probabilities are such, that the expected number of samples components is equal to the target rank $r$, but there is no guarantee. We modify the algorithm to always use exactly $r$ components, to allow for faster communication. We achieve this by repeating the sampling procedure until the number of selected components is $r$. This has no significant impact on the runtime performance.

\begin{algorithm}[H]
  \caption{Rank-$r$ Spectral-Atomo compression}\label{alg:spectral_atomo}
  \begin{algorithmic}[1]
  \Function{compress}{update matrix $M \in \R^{n \times m}$}
    \State $U, S, V \leftarrow \textsc{svd}(M)$.
      \Comment on CPU using Numpy, faster than \pytorch
    \State Compute Atomo probabilities $p_1 \ldots p_k$ from $S_{11},\ldots S_{kk}$.
      \Comment see \cite{wang2018atomo}.
    \State Sampling: include index $i$ independently with probability $p_i$.
    \State Repeat sampling until a set of $r$ indices $C$ is selected.
      \Comment our modification (see above)
    \State \Return $\left\{(U_{i:} \cdot S_{ii}/p_i, V_{i:}) \;|\; i \in C\right\}$ as two matrices $U^\prime \in \R^{n \times r}$ and $V^\prime \in \R^{m\times r}$.
  \EndFunction
  \Function{aggregate+decompress}{rank-$r$ approximations $(U^\prime_1,V^\prime_1) \ldots (U^\prime_W, V^\prime_W)$ for each worker}
    \State \Return $\sum_{i=1}^W U^\prime_i V_i^{\prime\top}$
      \Comment using \allgather\ in \nccl
  \EndFunction
  \end{algorithmic}
\end{algorithm}

The employed \allgather\ operation from \nccl\ is faster than aggregation using a parameter server using \gloo. We cannot implement a parameter server in \nccl\ due to lack of a `gather' operation.

\subsection{Best-approximation \powersgd}

This variant is the same as \powersgd\ (Algorithm~\ref{alg:powersgd}), but with more steps of subspace iteration, and without reuse of previous steps.
We find that 4 steps of subspace iterations (8 matrix multiplications) is enough to converge to the best low-rank approximation of gradient matrices, when measuring final test accuracy achieved by \powersgd.

\raggedbottom
\pagebreak

\section{Performance optimizations} \label{sec:optimizations}

Because we compare timings, we have aimed to optimize all compared optimizers to a similar level.
For sign-based methods, we used a publicly available C++ library by \cite{bernstein2019iclr} to efficiently pack the signs into bitmaps, an operation which is not supported by \pytorch\ natively.
For Atomo, we have benchmarked the SVD operation on the GPU and CPU, and chose the faster CPU implementation.
For all methods, we pack all gradient tensors into one flat buffer to reduce the number of communications.
Where possible, we overlay communication with computation.
Algorithms that do not support \allreduce\ are implemented using \nccl's \allgather, which is faster than a parameter server with \gloo.\footnote{`reduce'+`gather' (parameter server communication) with \gloo\ takes longer than \allgather\ with \nccl, as shown in Appendix~\ref{sec:cluster_specifications}. \nccl\ in \pytorch\ currently lacks support for a `gather' operator.}

\section{Learning rate tuning} \label{sec:lr_tuning}

For each task and each optimization algorithm without error feedback, learning rates were tuned separately. For algorithms based on error feedback with momentum, we use the learning rate tuned for SGD.

Learning rates are defined as rates for 1 worker, and scaled linearly with 5-epoch warmup to the number of workers (16 by default). We tune them in the 16-worker setting.

We determine the best learning rate by comparing test accuracy of one replication after running the full number of epochs.
We start training with 3 different learning rates, a factor 2 apart, based on commonly used rates for the optimizer, and if the best learning rate is either the lower or higher end, we extended the range.

For \cifar, the rates considered for SGD were [0.05, 0.1, 0.2], we chose 0.1. For rank-2 \atomo, we considered [0.025, 0.05, 0.1, 0.2] and chose 0.1. For \signum, we considered [2e-5, 5e-5, 1e-4, 2e-4] and chose 5e-5.

For \wikitext, the rates considered for SGD were [0.6, 1.25, 2.5, 5, 10], we chose 1.25. For \signum, we considered [2e-4, 1e-1, 5e-5, 1e-5, 1e-6], and chose 1e-5.

We have not tuned the momentum parameter or $L_2$, weight decay parameters or learning rate schedule for any experiment.

\end{document}